\documentclass{article} 
\usepackage{iclr2024_conference,times}

\usepackage{amsmath,amsfonts,bm}

\def\eqref#1{equation~\ref{#1}}

\def\1{\bm{1}}

\DeclareMathAlphabet{\mathsfit}{\encodingdefault}{\sfdefault}{m}{sl}
\SetMathAlphabet{\mathsfit}{bold}{\encodingdefault}{\sfdefault}{bx}{n}

\usepackage{hyperref}
\usepackage{url}

\title{Beyond Discriminative Regions: Saliency Maps as Alternatives to CAMs for Weakly Supervised Semantic Segmentation}

\iclrfinalcopy

\author{M. Maruf, Arka Daw, Amartya Dutta, Jie Bu \& Anuj Karpatne  \\
Department of Computer Science\\
Virginia Tech\\
Blacksburg, VA, USA \\
\texttt{\{marufm,darka,amartya,jayroxis,karpatne\}@vt.edu} \\
}

\begin{document}

\maketitle

\begin{abstract}
In recent years, several Weakly Supervised Semantic Segmentation (WS3) methods have been proposed that use class activation maps (CAMs) generated by a classifier to produce pseudo-ground truths for training segmentation models. While CAMs are good at highlighting discriminative regions (DR) of an image, they are known to disregard regions of the object that do not contribute to the classifier’s prediction, termed non-discriminative regions (NDR). In contrast, attribution methods such as saliency maps provide an alternative approach for assigning a score to every pixel based on its contribution to the classification prediction. This paper provides a comprehensive comparison between saliencies and CAMs for WS3. Our study includes multiple perspectives on understanding their similarities and dissimilarities. Moreover, we provide new evaluation metrics that perform a comprehensive assessment of WS3 performance of alternative methods w.r.t. CAMs. We demonstrate the effectiveness of saliencies in addressing the limitation of CAMs through our empirical studies on benchmark datasets. Furthermore, we propose random cropping as a stochastic aggregation technique that improves the performance of saliency, making it a strong alternative to CAM for WS3. 
\end{abstract}

\section{Introduction}

The goal in weakly supervised semantic segmentation (WS3) is to train segmentation models with coarse-scale supervision and without using pixel-level annotations. In recent years, several WS3 methods have been proposed that use image-level class labels to generate pseudo-ground truths for training segmentation models. Many of these methods employ localization methods such as Class Activation Maps (CAMs) \cite{zhou2016learning, selvaraju2016grad, chattopadhay2018grad}, generated from a pre-trained classifier, to guide the segmentation process.

CAMs are   \textbf{activation maps} generated by the last convolutional neural network (ConvNet) layer of the classification model, which is integrated with the class-specific weights of the final fully-connected layer to produce a score for every pixel. While Class Activation Maps (CAM) are good at highlighting discriminative regions (DRs) of an image (i.e., regions that contribute significantly to the classifier’s decision), CAMs are also known to ignore regions of the target object class that do not contribute to the classifier’s prediction, termed non-discriminative regions (NDRs). In particular, it has been shown that the activation maps in the final convolution layer only contain information relevant for classification, a phenomenon called \textit{information bottleneck} \cite{lee2021reducing}. As a result, CAMs are biased towards mostly finding DR while missing the NDR of the target object, which is equally important for the purpose of segmentation. A number of WS3 solutions thus require further processing of the CAM outputs to recover NDR for high segmentation accuracy \cite{lee2021reducing, lee2021anti, li2018tell, hou2018self,kolesnikov2016seed, araslanov2020single}.

In contrast to activation maps, \textbf{attribution maps} provide an alternative approach for assigning a score to every pixel based on its contribution to the final neural network prediction. The most commonly used attribution map is the gradient-based Saliency Maps \cite{simonyan2013deep}. The basic idea of saliency is to calculate the gradient of the target class score with respect to every pixel in the input image. Attribution maps are fundamentally distinct from activation maps obtained from the last layer of ConvNet models. However, despite the frequent use of attribution maps for neural network interpretability, their use in WS3 as an alternative to CAMs has largely been unexplored.  

With the advancement of vision transformers achieving state-of-the-art (SOTA) performance on many computer vision tasks \cite{han2022survey}, extending CAMs to work with non-ConvNet-based classifiers is a non-trivial exercise. In contrast, gradient-based Saliency maps can be applied to any classifier with differentiable layers, rendering them as a universal solution for WS3 tasks. Moreover, Saliency maps inherently provide a solution to the deficiencies of CAM-based approaches as explored in this work. Although the limitations of CAMs have been well-known in the WS3 research community and all SOTA methods in WS3 provide solutions to mitigate the deficiencies of CAMs, they lack in providing deeper insights on how saliencies can be used as an alternative to CAM for WS3.

Our goal in this paper is to provide a comprehensive study of the comparison between CAMs and Saliecies for WS3. It is important to mention that our goal is not to achieve SOTA performance for WS3, but rather to provide novel insights into the potential of saliencies and their variations in addressing the limitations of CAMs. Our contributions are outlined below:

\begin{itemize}
    \item We offer multiple perspectives to understand the similarities and differences between CAMs and Saliencies. Section 3 delves into these perspectives, serving as a ``bridge'' in the analysis of CAMs and saliencies.
    \item We provide new evaluation metrics to measure WS3 performance, which are specifically designed to complement existing metrics such as mIoU in quantifying the deficiencies of CAMs and evaluating the effectiveness of alternate techniques w.r.t. CAMs.  The proposed evaluation metrics are detailed in Section 4.
    \item We demonstrate the effectiveness of saliencies in addressing the limitation of CAM through our empirical studies on the PASCAL VOC, COCO, and MNIST datasets, as detailed in  Section 5.
    \item We identify the limitations of saliency maps for WS3 and propose different variations of stochastic aggregation methods to fix these limitations. Specifically, we propose a random cropping approach for stochastic aggregation that disintegrates the spatial structure of input images as compared to injecting spatially invariant noise. While random cropping is a common data augmentation technique, its application as a stochastic aggregation method in this work is novel. Additional insights regarding stochastic aggregation of saliencies are presented in Sections 6 and 7.
\end{itemize}

\section{Fundamental Concepts and Definitions}
\label{sec:cams_and_saliencies}

\subsection{Class Activation Maps}
The Class Activation Maps (CAMs) are based on convolutional neural networks with a global average pooling (GAP) layer applied before the final layer. Formally, let the classifier be parameterized by $\theta=\{\theta_f, \mathbf{w}\}$, where $f(.;\theta_f)$ is the feature extractor network prior to the GAP layer and $\mathbf{w}$ is the set of weights of the final classification layer. The CAM of the $c$-th class for an image $\mathbf{I}$ can be obtained as follows:
\begin{equation}
    \text{CAM}_c(\mathbf{I}; \theta) = \frac{\mathbf{w}_{c}^{\text{T}} \mathbf{A}}{\max \mathbf{w}_{c}^{\text{T}} \mathbf{A}}
\end{equation}
where $\mathbf{A}=f(\mathbf{I};\theta_f)$ is the activation map, $\mathbf{w}_c \in \mathbf{w}$ is $c$-th class weight, and $\max(.)$ is the maximum value over all pixels in $\mathbf{I}$ for normalization. 

\subsubsection{Limitations of CAMs}
CAMs produce coarse-scale localizations of objects because the activation maps of the final convolutional layer have significantly lower resolution compared to the input image. Additionally, the final activation maps show high values for only a subset of regions of the target object that are discriminative for the classification task, while disregarding regions that do not impact the accuracy of classification. Thus, CAMs in their raw form without supplementary post-processing, are unsuitable for training segmentation models.


\subsubsection{Discriminative and Non-Discriminative Regions} 
\emph{Discriminative regions (DRs)} are those regions of the ground-truth object that are crucial for the classification model to predict the class label of the image accurately. In contrast, \emph{non-discriminative regions (NDRs)} are those regions of the ground-truth that are still important for segmenting the object but do not significantly impact the model's accuracy upon removal. 
We formally define DR and NDR based on the CAM outputs as follows:

\begin{definition}[DR and NDR]
\label{def:dr_ndr}
The discriminative region (DR) and non-discriminative region (NDR) for the $c$-th class of an image $\mathbf{I}$ can be defined for every pixel $(i, j)$ belonging to the $c$-th class ground-truth segmentation $\mathcal{S}^c_{GT}$ as follows:
\begin{align}
    \text{DR}_c(i, j) &= \mathbb{I}(\text{CAM}_c(i, j) \geq \tau_{cam})\\
    \text{NDR}_c(i, j) &= \mathbb{I}(\text{CAM}_c(i, j) < \tau_{cam})
\end{align}
where $\tau_{cam}$ represents a threshold applied to the CAM to obtain the segmentation of the object class and $\mathbb{I}(.)$ is the indicator function. While the optimal threshold  may differ for each image, we adopted the common practice of using a global threshold $(\tau_{cam}=0.25)$ for defining DR and NDR throughout this paper. Note that DRs and NDRs are a partitioning of the ground-truth mask $\mathcal{S}_{GT}^c$ based on CAM scores. 

\end{definition}



\subsection{Saliency Maps}
Saliency maps are attribution maps that assign a score to every image pixel representing its contribution to the final classifier prediction. They are frequently employed as a tool to enhance model interpretability. Formally, the saliency map (SM) of the $c$-th class for image $\mathbf{I}$ can be defined as:

\begin{align}
\label{eq:sal_map}
    \text{SM}_c(\mathbf{I}, \theta) &= \Big|\frac{\partial{S_c}}{\partial{\mathbf{I}}} \Big| = \Big|\mathbf{w}_{c}^{\text{T}} \frac{\partial{\text{GAP}(\mathbf{A})}}{\partial{\mathbf{I}}} \Big| 
\end{align}

where $S_c = \mathbf{w}_{c}^{\text{T}} \text{GAP}(\mathbf{A}) + b_c$ is the score for the $c$-th class, and $b_c \in \mathbf{w}$ is the bias term. For a multi-channel image, saliency maps are computed by taking a maximum of the gradient values across the channels.


\begin{definition}[HSR and LSR]
\label{def:hsr}
The high saliency region (HSR) and low saliency region (LSR) for the $c$-th class of an image $\mathbf{I}$ can be defined for every pixel $(i, j)$ belonging to the $c$-th class ground-truth segmentation $\mathcal{S}^c_{GT}$ using a threshold $\tau_{sm}$ specific to saliency maps as follows:
\begin{align}
    \text{HSR}_c(i, j) &= \mathbb{I}(\text{SM}_c (i, j) \geq \tau_{sm}) \\
    \text{LSR}_c(i, j) &= \mathbb{I}(\text{SM}_c (i, j) < \tau_{sm})
\end{align}
\end{definition}

Just like DRs and NDRs, the HSRs and LSRs are an alternate partitioning of $\mathcal{S}_{GT}^c$ based on SM score.

\begin{figure}[t]
\centering
\begin{subfigure}{0.25\linewidth}
   \includegraphics[width=\linewidth]{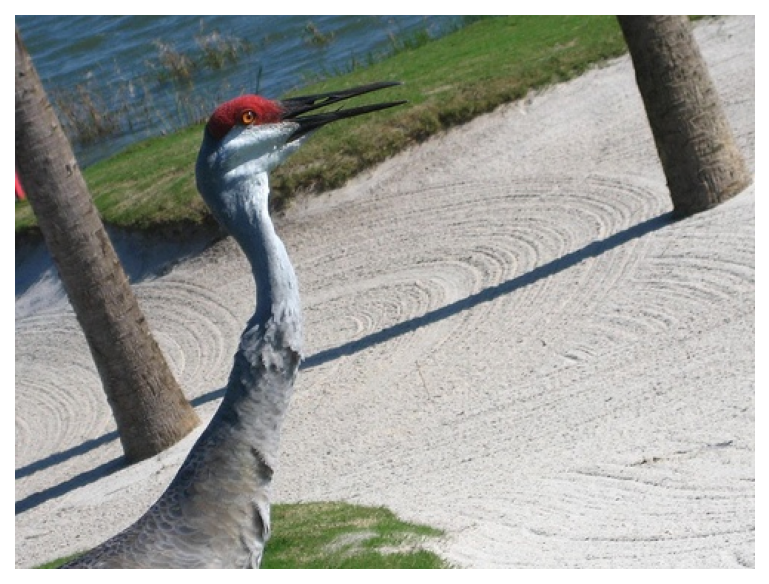}
   \caption{Original Image}
   \label{subfig:original}
\end{subfigure}
\begin{subfigure}{0.25\linewidth}
   \includegraphics[width=\linewidth]{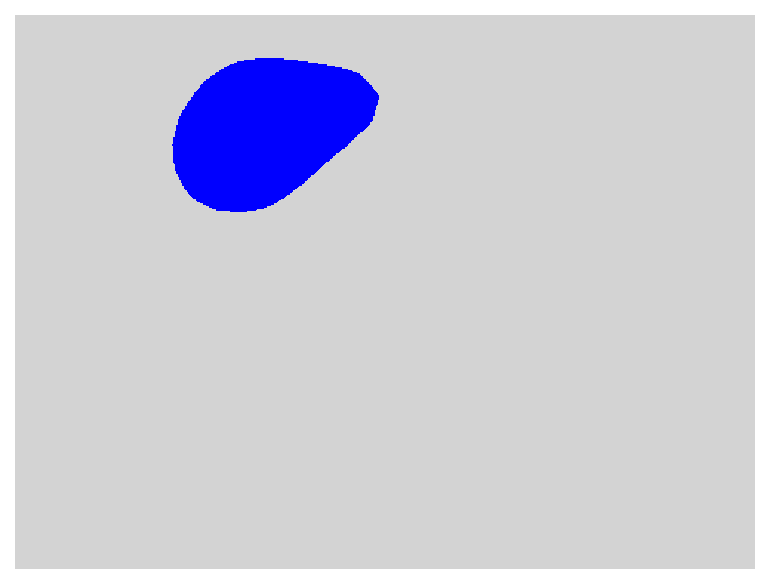}
   \caption{CAM}
   \label{subfig:cam}
\end{subfigure}
\begin{subfigure}{0.25\linewidth}
   \includegraphics[width=\linewidth]{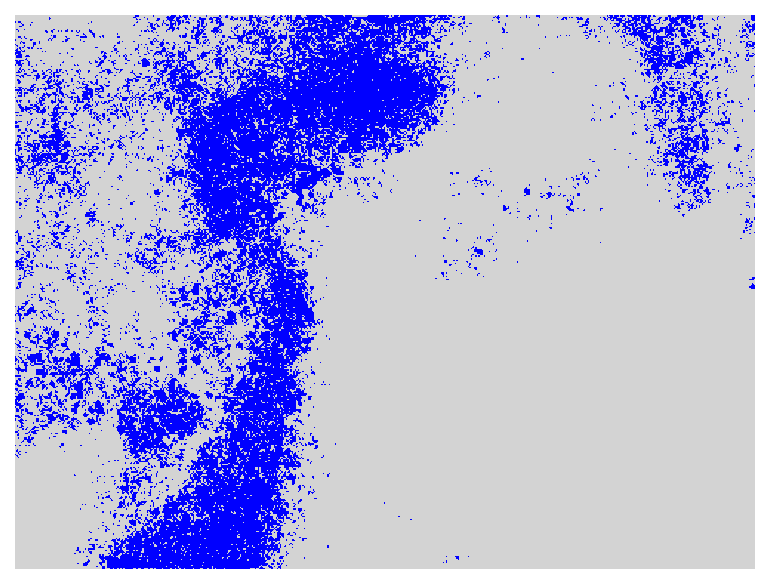}
   \caption{Saliency Map}
   \label{subfig:saliency}
\end{subfigure}

\begin{subfigure}{0.32\linewidth}
   \includegraphics[width=\linewidth]{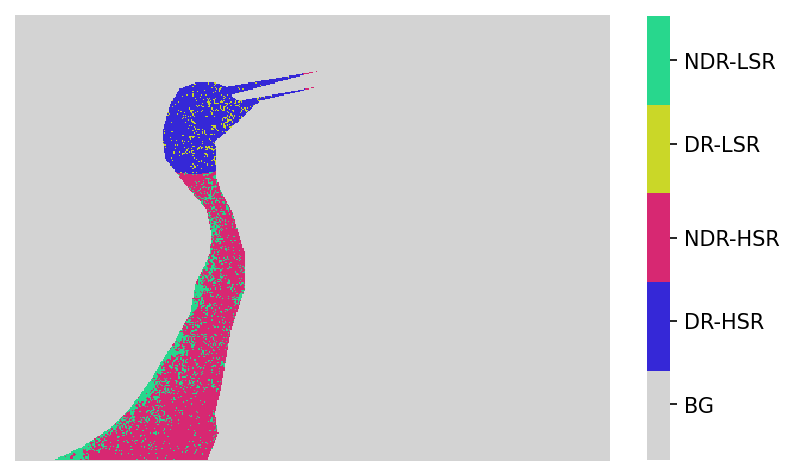}
   \caption{CAM vs. SM - Image}
   \label{subfig:segmap}
\end{subfigure}
\begin{subfigure}{0.3\linewidth}
   \includegraphics[width=\linewidth]{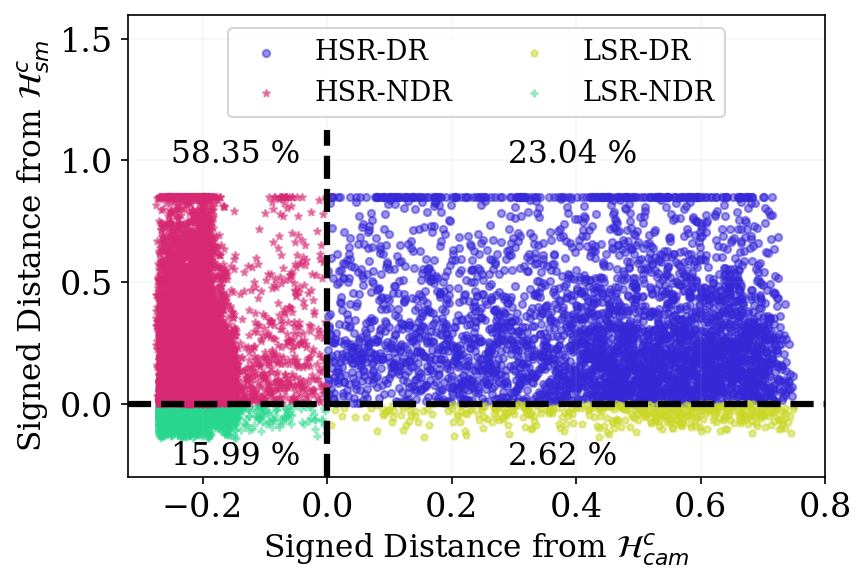}
   \caption{CAM vs. SM - Distance}
   \label{subfig:dist_quad}
\end{subfigure}
\caption{A visual comparison of CAMs and saliency maps (SMs) for a representative image from the VOC12 dataset.}
\label{fig:hyperplane}
\end{figure}


\section{Comparing CAMs and Saliency Maps}



\subsection{A Visual Comparison Using Hyperplanes}

While CAMs and saliency maps differ in many respects, they also exhibit several similarities. We offer a novel viewpoint of comparing CAMs and SMs from the lens of CAM and SM hyperplanes. First, we  define two $k$-dimensional Hilbert spaces (where $k$ is the number of channels in the activation map): $\mathcal{A}$ for the activations of images and $\mathcal{A'}$ for the gradients of the GAP layer w.r.t. the image. Formally, for an arbitrary image $\mathbf{I}$, let the activation at any pixel $\mathbf{A}_{(i, j)} \in \mathcal{A}$, and the gradient of the GAP layer $\frac{\partial \text{GAP}(\mathbf{A})}{\partial \mathbf{I}}|_{(i, j)} \in \mathcal{A'}$. 


\begin{definition}[$c$-th class CAM hyperplane]
\label{def:cam_hyperplane}
For every image $\mathbf{I}$, let $\mathcal{H}^c_{cam}$ be the following hyperplane in $\mathcal{A}$:
\begin{equation}
   \mathcal{H}^c_{cam} \: : \: \frac{\mathbf{w}_{c}^{\text{T}}}{Z} \mathbf{a} - \tau_{cam} = 0
\end{equation}
where $\tau_{cam}$ is the CAM threshold, $\mathbf{w}_{c} \in \mathbf{w}$ is the weight for the $c$-th class, and $Z = \max \mathbf{w}_{c}^{\text{T}} \mathbf{A}$ is a normalization factor depending on $\mathbf{I}$. Note that $Z$ changes for every image and is equivalent to having a variable intercept term for the CAM hyperplane  but with a fixed slope $\mathbf{w}_{c}$ for every image.
\end{definition}

\begin{remark}
\label{remark:dr}
If a  point $\mathbf{a} \in \mathcal{A}$ corresponding to a ground-truth pixel lies above $\mathcal{H}^c_{cam}$, i.e., $\mathbf{w}_{c}^{\text{T}} \mathbf{a}/Z - \tau_{cam} \geq 0$, then the pixel belongs to DR; otherwise, it belongs to NDR.
\end{remark}
See Appendix for proof. This remark states that any arbitrary pixel $(i,j) \in \mathcal{S}^c_{GT}$ will belong to the DR or NDR depending on which side of the CAM hyperplane it lies. In other words, as long as $\mathbf{w_c}$ and $\tau_{cam}$ are fixed, the DR and NDR of the $c$-th class for any image $\mathbf{I}$ are separated by its CAM hyperplane $\mathcal{H}^c_{cam}$.

\begin{definition}[$c$-th class SM parallel-hyperplane]
\label{def:sm_hyperplane}
Let $\mathcal{H}^c_{sm}$ be the following set of two parallel hyperplanes in $\mathcal{A'}$:
\begin{align}
    \mathcal{H}^c_{sm} \: : \: |\mathbf{w}_{c}^{\text{T}} \mathbf{a'}|- \tau_{sm} = 0 
\end{align}
where $\tau_{sm}$ is the saliency map threshold and $\mathbf{a'} \in \mathcal{A'}$ is the gradient of the GAP layer w.r.t. image at any arbitrary pixel.
\end{definition}

\begin{remark}
\label{remark:hsr}
If a point $\mathbf{a'}$ corresponding to a ground-truth pixel lies on the outer sides of $\mathcal{H}^c_{sm}$, i.e., $|\mathbf{w}_{c}^{\text{T}} \mathbf{a'}| - \tau_{sm} \geq 0$, then the point belongs to HSR; otherwise, it belongs to LSR.
\end{remark}

See appendix for proof. Similar to the DR/NDR for CAMs, the HSR/LSR are separated by SM parallel-hyperplanes. Furthermore, the slope of both CAM and SM hyperplanes are the same: $\mathbf{w_c}$. However, the important distinction is that for CAMs, the DR/NDR depends on the values of the activation map $\mathbf{A}_{(i, j)}$, while for SMs, the HSR/LSR depends on the gradient $\frac{\partial \text{GAP}(\mathbf{A})}{\partial \mathbf{I}}|_{(i, j)}$. A ground-truth pixel may thus belong to DR or NDR and HSR or LSR depending on the value of its activations and gradient of GAP layer, respectively.

In Figure \ref{fig:hyperplane}, we visually compare CAMs and SMs for a representative image from the VOC12 dataset. From this comparison, we observe that the CAM (see Figure \ref{subfig:cam}) predominantly highlights the DR of the bird class such as its head — a crucial feature for classification. As a result, NDRs such as the bird's body are sparingly covered by the CAM. In contrast, the saliency map (see Figure \ref{subfig:saliency})  for the same image covers most regions of the target bird class, albeit with some noisy representation of the background class too. To provide a comprehensive visualization of how HSRs in saliency maps can potentially recover NDRs, we present a scatterplot in Figure \ref{subfig:dist_quad} comparing the signed distances of each pixel $(i, j) \in \mathcal{S}^{bird}_{GT}$ from the CAM and SM hyperplanes, namely, $\mathcal{H}^{bird}_{cam}$ and $\mathcal{H}^{bird}_{sm}$. Notably, the HSRs successfully recover a substantial portion of DRs, labeled as HSR-DR (blue). A minor segment of the DRs ($2.62 \%$ of GT) is missed by SMs, termed LSR-DR (yellow). Nonetheless, SMs are proficient in recovering $55.32 \%$ of the GT regions originally classified as NDR, labeled as HSR-NDR (maroon). Yet, both SMs and CAMs fall short in capturing the LSR-NDR region, which constitutes $15.99 \%$ of the GT (green). The color-coded segmentation map for these four distinct regions are presented in Figure \ref{subfig:segmap}, thereby showing the potential of saliency maps in addressing the limitations of CAMs in recovering NDRs.


\subsection{Perspective from Contribution Windows}

Next, we present another novel viewpoint of comparing saliencies and CAMs from the perspective of \textit{contribution windows}---a concept innate to the architecture of convolutional neural networks (ConvNets). Note that the tendency of CAMs to only focus on DRs can be understood using the \textit{information bottleneck} principle proposed in \cite{lee2021reducing}---every layer of a neural network filters or ``funnels in'' information about inputs and as a result only task-specific information is retained at the outputs. While this information bottleneck exists in the forward propagation of ConvNets, the reverse phenomenon happens during backpropagation when information ``funnels out" from the activation maps to the input image. This phenomenon can be described using the {contribution window} of an input pixel on the activation maps, defined as follows.

\begin{definition}[Contribution Window]
Let's consider a ConvNet with $N$ layers, where every layer $l$ performs a 2D convolution using an $F \times F$ kernel denoted as $\mathbf{K}_l$, to compute activation $\mathbf{A}_l = \text{Conv2D}(\mathbf{A}_{l-1}, \mathbf{K}_l)$. The contribution window at layer $l$ of a pixel in the input image can then be defined as the region in $\mathbf{A}_l$ that affects (or contributes to) the gradients of $\mathbf{A}_{l}$ w.r.t. the input pixel. 
\end{definition}

\begin{figure}[htb]  
    \centering  
    \includegraphics[width=0.55\linewidth]{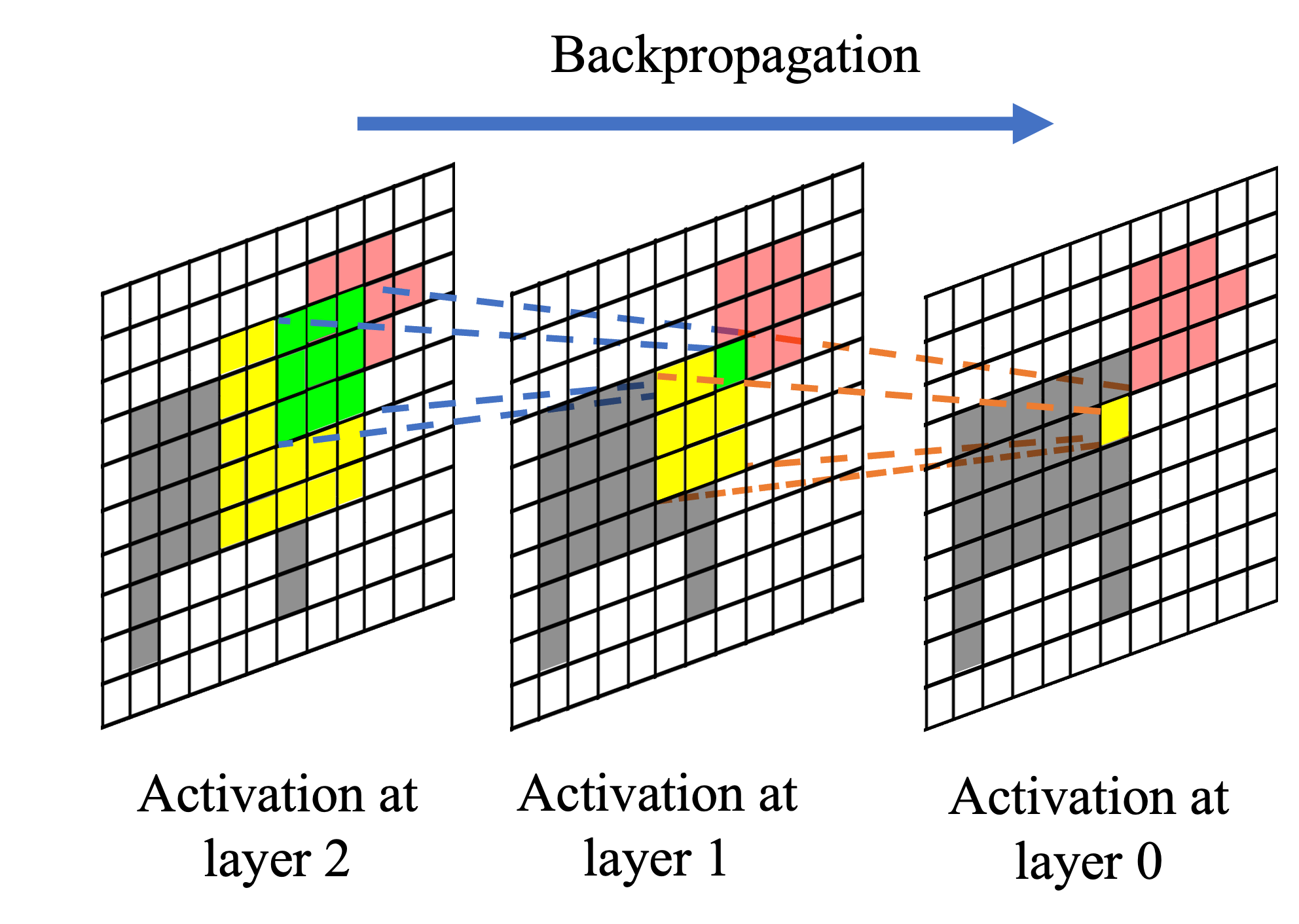}
    \caption{A schematic of ``contribution window'' demonstrating how the gradients at layer $l-1$ is affected by the gradients from the contribution window of layer $l$.}
    \label{subfig:contri_win}  
\end{figure}

This concept is illustrated in Figure \ref{subfig:contri_win}, where the contribution window is highlighted in yellow at every layer for an example yellow pixel at layer 0. The contribution window can be viewed as the reverse concept of ``receptive fields" defined for the forward pass of ConvNets. Indeed, since the gradient of the forward convolution $\mathbf{K}_l$ is also a convolution with a rotated kernel \cite{Kafuna_2016}, the receptive field of the backward convolution during gradient computation becomes the concept of contribution window. We can show that all activations at layer $l$ in the contribution window of an input pixel can affect its gradient. 

Now, let us consider pixels that have $0$ activations across all channels in the final layer shown in grey in Figure \ref{subfig:contri_win}. By design, such \textit{non-activated pixels} will register $0$ CAM scores. We want to analyze if it is possible for a non-activated pixel (yellow) to show non-zero gradients (and thus saliencies) in the input image. Assuming we use activation functions $f(z)$ that are $0$ when $z \leq 0$, we can show that this depends on whether the contribution window of the pixel contains any \textit{activated pixel} with non-zero activations at the final layer, shown in red. In fact, we can show that if the contribution window size of a non-activated pixel is smaller than its distance from an activated pixel, it will have 0 gradients. However, this is practically not likely as the contribution window size generally grows linearly with the depth of ConvNets. An exception is when we use $1 \times 1$ kernels. Through empirical evidence provided in section \ref{sec:contr_win_exp}, we can establish that as the contribution window expands (achieved by increasing the $F \times F$ kernel size), saliencies can progressively encompass more NDRs, thus directly addressing the limitations of CAMs.

\section{Experimental Setup \& Evaluation Metrics}
\label{sec:setup_eval_metrics}

\subsection{Experimental Setup:}
\label{sec:experimental_setup}
Following the common practice in WS3, in this paper, we compared  different approaches quantitatively and qualitatively by conducting experiments on MNIST, PASCAL VOC '12, and MS COCO '14 datasets. We also utilized two types of classification models based on ResNet50 architecture: i) ``model-org", which is simply fine-tuned on the corresponding dataset, and ii) ``model-pert", which is fine-tuned with additional noise perturbation. 

\subsection{Evaluation Metrics:}
\label{sec:evaluation_metrics}
To assess the quality of the segmentation maps, \emph{mean intersection over union (mIoU)} is a widely used metric in WS3 literature. mIoU measures the ratio of correct prediction (intersection) over the union of predictions and ground truths, averaged across all classes, including background class. Notably, mIoU provides an unbiased estimate of the segmentation performance; however, it fails to provide insights about the coverage of NDRs and DRs. Given the limitation of CAMs not being able to identify NDRs, it becomes crucial to measure how effective alternative WS3 techniques (e.g., saliencies) are at addressing the deficiencies of CAMs. This warrants the need for novel evaluation metrics focusing on the DRs and NDRs.

In this paper, we introduce the following three novel evaluation metrics: NDR-Recall, DR-Recall, and Foreground Precision (FG-Prec). \textbf{DR-Recall} is the ratio of correct DR prediction over the ground-truth DR and can be formally defined as: $\text{DR-Recall} = |\text{TP}(P, DR_{GT})|/(|\text{TP}(P, DR_{GT})| + |\text{FN}(P, DR_{GT})|)$, where $P$ denotes the segmentation prediction, $DR_{GT}$ denotes the ground-truth DR area, and $|\text{TP}|$ and $|\text{FN}|$ denote the count of true positives and false negatives over the DR region. As mentioned in Section \ref{def:dr_ndr}, we define ground truth DR ($DR_{GT}$) and NDR ($NDR_{GT}$) by employing a global threshold $(\tau_{cam} = 0.25)$ on the CAM prediction and then taking its overlap with the ground-truth segmentation mask. In a similar manner, we compute \textbf{NDR-Recall} for a given segmentation prediction ($P$) and the corresponding ground-truth NDR region ($NDR_{GT}$).
Apart from these two metrics, we also compute the \textbf{Foreground-Precision} of different target-classes as an additional metric, which can be defined as the ratio of correct foreground prediction over the total foreground prediction.
Note that our proposed metrics are defined to analyze the deficiencies of CAM and hence, are biased only if we are evaluating CAMs just by themselves (e.g., CAMs would show low NDR Recall value by definition). However, these metrics are unbiased if the goal is to measure how well alternative WS3 techniques (e.g., saliencies) fix the shortcomings of CAMs.

\section{Quantitative Comparison: CAM/Saliency}

\subsection{Effect of Contribution Window}
\label{sec:contr_win_exp}

To empirically demonstrate the effect of contribution window on the recovery of NDRs, we utilize a 5-layer ConvNet architecture where each layer employs an $F \times F$ kernel, followed by ReLU activation. 
We apply sufficient zero padding to ensure that the spatial dimension of the activations in each layer is equal to that of the input image. 
Different models with varying kernel sizes were then trained on the MNIST Segmentation dataset.

\begin{figure}[htb]  
    \centering  
    \includegraphics[width=0.4\linewidth]{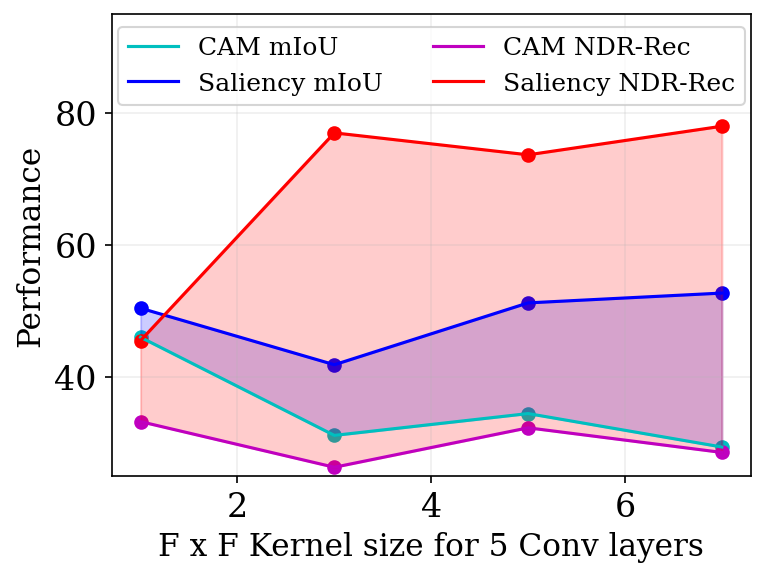}
    \caption{Effect of Contribution Window on NDR-Recall and mIoU for MNIST Dataset.}
    \label{subfig:contri_win_vs_perf}  
\end{figure}

The results for CAM and Saliency, in terms of mIoU and NDR-Recall, are presented in Figure \ref{subfig:contri_win_vs_perf}. The $F \times F$ kernel size correlates with the size of the contribution window for the backpropagated gradients. Notably, when the contribution window is $1 \times 1$, the performance of CAMs and Saliencies is quite comparable. However, differences in performance become more prominent (larger red and blue shaded regions) as the contribution window size increases. With an expanding contribution window, saliencies are capable of recovering more pixels that have high gradients and low ($\approx 0$) activations, effectively capturing a larger proportion of NDR. This, in turn, leads to a gradual increase in NDR-Recall until saturation is achieved. Further discussion of this experiment can be found in the Appendix.



\subsection{Comparing NDR Recovery}
\begin{table}[t]
\renewcommand{\arraystretch}{1.2}
\centering
\setlength\tabcolsep{3pt} 
\fontsize{9pt}{9}\selectfont
\begin{tabular}{cl|cccc}
\hline
\textbf{Method} & \multicolumn{1}{c|}{\textbf{B/G Resolve}} & \textbf{mIoU} & \textbf{FG-Prec} & \textbf{DR-Recall} & \textbf{NDR-Recall} \\ \hline
CAM & Basic & 43.7 & 56.1 & \textbf{93.8} & 43.7 \\ \hline
\multirow{3}{*}{Saliency} & Basic & 37.7 & 45.9 & 75.4 & 55.6 \\
 & Smooth & 44.0 & 52.2 & 84.3 & 60.0 \\
 & Superpixel & \textbf{49.0} & \textbf{60.0} & 80.9 & \textbf{61.8} \\ \hline
\end{tabular}

\caption{Quantitative comparison of CAM and Saliency on VOC dataset in terms of mIoU, Foreground Precision, and DR-/NDR-Recall.}
    \label{tab:quant_compare_cam_sal}

\end{table}


Table \ref{tab:quant_compare_cam_sal} presents quantitative evaluation of CAMs and saliencies on the PASCAL VOC dataset using different methods for background resolve (see Appendix for details). We compare the best-segmented map produced by each method by varying the global threshold of $\tau_{cam}$ and $\tau_{sm}$ from $0.01$ to $0.50$ and selecting the segmented map with the highest mIoU. The ``basic background resolve'' row of Table \ref{tab:quant_compare_cam_sal} shows that  saliency map outperforms CAM in finding non-discriminative regions, as indicated by its higher NDR-Recall score. However, CAM outperforms the saliency maps in terms of mIoU, FG-precision, and DR-Recall, likely due to the noisy and scattered nature of saliency maps. This motivates further exploration of opportunities to improve the quality of saliency maps.

\subsection{Improving Saliencies with Simple Post-processing}
\label{sec:saliency_post_proc}
We first explore if simple post-processing methods such as \textbf{kernel smoothing background resolve} and \textbf{Superpixel-based background resolve} can improve SM performance. 
Kernel Smoothing smooths the gradients of the saliencies by applying a Gaussian kernel, while superpixel-based smoothing assigns a label to each superpixel, which effectively mitigates the noisiness and scatteredness that may be present in saliency maps. See Appendix for details of these post-processing approaches.
Table \ref{tab:quant_compare_cam_sal} presents their results as `Smooth' and `Superpixel' background Resolve. Both approaches outperform basic background resolve results in terms of mIoU, FG-Precision, DR-Recall, and NDR-Recall. Superpixel-based saliency maps demonstrate significant improvement over CAM in terms of mIoU and NDR-Recall; however, CAM outperforms all saliency methods in finding discriminative regions, as indicated by its higher DR-Recall score. It is worth mentioning that superpixel-based background resolve is not scalable for larger datasets.  To this end, we need to explore saliencies where the smoothing can be integrated inherently without additional computational overheads.

\begin{table}[t]
\renewcommand{\arraystretch}{1.3}
\setlength\tabcolsep{3.5pt} 
\fontsize{9pt}{9}\selectfont
\centering
\begin{tabular}{l|l|l|cccc}
\hline
\textbf{Model} & \textbf{Method} & \textbf{BG-Res} & \multicolumn{1}{l}{\textbf{\:\: mIoU}} & \multicolumn{1}{l}{\begin{tabular}[c]{@{}l@{}}\textbf{FG-Prec}\end{tabular}} & \multicolumn{1}{l}{\begin{tabular}[c]{@{}l@{}}\textbf{DR-Rec} \end{tabular}} & \multicolumn{1}{l}{\begin{tabular}[c]{@{}l@{}}\textbf{NDR-Rec} \end{tabular}} \\ \hline
\multirow{3}{*}{org} & \multirow{3}{*}{\begin{tabular}[c]{@{}l@{}}Smooth-\\ Grad\end{tabular}} & Basic & 38.6 (+0.9) & 47.1 (+1.2) & 82.0 (+6.6) & 51.7 (-3.9) \\
 &  & Smooth & 37.5 (-6.5) & 47.1 (-5.1) & 79.2 (-5.1) & 48.3 (-11.7) \\
 &  & Superpix & 41.0 (-8.0) & 52.2 (-7.8) & 77.0 (-3.9) & 52.1 (-9.7) \\ \hline
\multirow{3}{*}{\begin{tabular}[c]{@{}l@{}}pert-\\ gauss\end{tabular}} & \multirow{3}{*}{\begin{tabular}[c]{@{}l@{}}Smooth-\\ Grad\end{tabular}} & Basic & 45.3 (+7.6) & 54.9 (+9.0) & 87.4 (+12.0) & 55.9 (+0.3) \\
 &  & Smooth & 44.8 (+0.8) & 54.1 (+1.9) & \textbf{87.5 (+3.2)} & 56.8 (-3.2) \\
 &  & Superpix & 48.1 (-0.9) & \textbf{57.4 (-2.6)} & 86.4 (+5.5) & 62.9 (+1.1) \\ \hline
\multirow{3}{*}{org} & \multirow{3}{*}{\begin{tabular}[c]{@{}l@{}}Binary-\\ Mask\end{tabular}} & Basic & 41.2 (+3.5) & 51.3 (+5.4) & 79.9 (+4.5) & 53.6 (-2.0) \\
 &  & Smooth & 43.4 (-0.6) & 53.5 (+1.3) & 84.7 (+0.4) & 53.9 (-6.1) \\
 &  & Superpix & 47.3 (-1.7) & 57.0 (-3.0) & 84.8 (+3.9) & 62.0 (+0.2) \\ \hline
\multirow{3}{*}{\begin{tabular}[c]{@{}l@{}}pert-\\ binary\end{tabular}} & \multirow{3}{*}{\begin{tabular}[c]{@{}l@{}}Binary-\\ Mask\end{tabular}} & Basic & 42.4 (+4.7) & 52.9 (+7.0) & 78.7 (+3.3) & 55.8 (+0.2) \\
 &  & Smooth & 44.9 (+0.9) & 54.8 (+2.6) & 84.8 (+0.5) & 57.2 (-2.8) \\
 &  & Superpix & \textbf{48.9 (-0.1)} & 56.8 (-3.2) & 86.2 (+5.3) & \textbf{68.0 (+6.2)} \\ \hline
\end{tabular}

\caption{Quantitative comparison of SmoothGrad and BinaryMask in terms of mIoU, FG-Precision, DR-/ NDR-Recall for different fine-tuned models on VOC dataset. The difference between the aggregated saliency performance and the vanilla saliency performance is shown in parentheses. A positive value denotes an increase in performance; whereas a negative value denotes a decrease in performance for aggregated saliencies. }
\label{tab:stochastic_quant}
\end{table}

\section{Stochastic Aggregation of Saliencies}
\label{sec:stochastic_saliency}

To reduce the noisiness of saliencies, \cite{smilkov2017smoothgrad} proposed a stochastic aggregation-based method for saliency maps, named \textbf{SmoothGrad}, where Gaussian noise is added to the input image for smoothing saliencies.
In this paper, we explored another variation of input noise perturbation, namely \textbf{BinaryMask}, where we multiply the image by a binary mask instead of adding Gaussian noise to the input image. The amount of perturbation for SmoothGrad is controlled by standard deviation of Gaussian noise, whereas for BinaryMask, the probability of each pixel in the mask being 1 controls the perturbation magnitude. See Appendix for additional details on these methods. ``\textit{Model-pert-binary}" and ``\textit{Model-pert-gaussian}" are the two finetuned classifiers augmented by binary and Gaussian noise, respectively.

\subsection{Smoothing Saliencies by Injecting Noise}
Table \ref{tab:stochastic_quant} compares results of saliency with different stochastic aggregation methods like SmoothGrad and BinaryMask. The change in performance from the basic or vanilla saliencies (without stochastic aggregation) is shown in parentheses; a positive percentage denotes improvement and a negative percentage denotes degradation. Saliencies from the classification models perturbed with similar noise (\textit{model-pert-gaussian} for \textbf{SmoothGrad} and \textit{model-pert-binary} for \textbf{BinaryMask}) perform better than the saliencies generated by the original model. According to \cite{bishop1995training}, adding noise during training is a common regularization technique that results in denoising. The additive effect of adding noise during training and inferring with noise yields the best saliency map.



Although adding noise may make the saliency maps smoother, with increasing noise, the saliency maps may become unstable and the mIoU performance may gradually drop with excessive noise. A detailed analysis of the sensitivity of our experiments to noise is provided in Appendix. Also note that the classification model needs to be fine-tuned with similar noise for these stochastic perturbations techniques to produce smoothened saliencies. This additional fine-tuning could be an expensive process, and further motivates us to explore alternate aggregation methods that do not involve additional fine-tuning steps.

\section{Stochastic Aggregation Through Cropping}
\label{sec:stochastic_random_crop}


Random cropping is commonly used as a data augmentation technique to increase the variety of training data by cropping random regions of the input image to a specific size. 
One of the advantages of random cropping is that it generates input samples that follow the input data distribution, since all the crops are basically part of the input image. 
In this section, we utilize random cropping as a stochastic aggregation technique to improve the performance of saliencies.

\subsection{Disintegrating the Spatial Structure of Images using Random Cropping}

Random cropping can also be viewed as a perturbation technique where the individual crops disintegrate the spatial structure of the input image. We treat random cropping as a spatial perturbation and generate a saliency map by stochastically aggregating the saliency maps of the individual cropped images.
We define this spatial perturbation-based aggregation as follows: $\tilde{\text{SM}}_c(\mathbf{I}) = \frac{1}{n} \sum_{i=1}^{n} w_i\text{SM}_c(\tilde{\mathbf{I}}_i)$, where $\tilde{\mathbf{I}}_i = f_{pert}(\mathbf{I})$, $\mathbf{I}$ corresponds to the input image, $\tilde{\mathbf{I}}_i$ denotes the individual crops, and $f_{pert}(.)$ denotes the spatial perturbation function, which is random cropping for this experiment. $\text{SM}_c(.)$ is the (basic) saliency map and $\tilde{\text{SM}}_c$ corresponds to the final aggregated saliency, and $w_i$ denotes the weight of each of the individual crop saliencies. For our experiments, we choose $w_i = \sigma(S_c(\tilde{\mathbf{I}}_i))$, where $S_c(\tilde{\mathbf{I}}_i)$ is the classification score of the crop $\tilde{\mathbf{I}}_i$, and $\sigma(.)$ is the sigmoid activation function. 


First row of Table \ref{tab:cropping_quant} shows the performance of random cropping as a stochastic aggregation method, where we can see that it performs better than saliencies in terms of mIoU, FG-Precision, and DR-/NDR- Recall for all the background resolve approaches (difference in performance of random crop and saliencies are provided in parentheses). We can achieve as high as 50.4 mIoU using random crop-based aggregated saliencies with superpixel-based background resolve. Notably, random cropping-based aggregated saliencies employ the ``\textit{Model-org}'' classifier to compute the saliencies, showing that random cropping does not require the classifier to be finetuned on additional perturbations to perform well.

\begin{table}[h]
\renewcommand{\arraystretch}{1.2}
\setlength\tabcolsep{3pt} 
\fontsize{9pt}{9}\selectfont
\centering
\begin{tabular}{l|l|cccc}
\hline
\textbf{Method} & \textbf{BG-Res} & \multicolumn{1}{l}{\textbf{mIoU}} & \multicolumn{1}{l}{\begin{tabular}[c]{@{}l@{}}\textbf{FG-Precision}\end{tabular}} & \multicolumn{1}{l}{\begin{tabular}[c]{@{}l@{}}\textbf{DR-Recall}\end{tabular}} & \multicolumn{1}{l}{\begin{tabular}[c]{@{}l@{}}\textbf{NDR-Recall} \end{tabular}} \\ \hline
\multirow{3}{*}{\begin{tabular}[c]{@{}l@{}}Random\\ Crop\end{tabular}} & Basic & 44.6 (+6.9) & 53.6 (+7.7) & 84.2 (+8.8) & 59.4 (+3.8) \\
 & Smooth & 46.2 (+2.2) & 56.6 (+4.4) & \textbf{84.4 (+0.1)} & 57.5 (-2.5) \\
 & Superpix & 50.4 (+1.4) & \textbf{61.7 (+1.7)} & 82.6 (+1.7) & \textbf{61.7 (-0.1)} \\ \hline
\multirow{3}{*}{\begin{tabular}[c]{@{}l@{}}Random\\ Patch\end{tabular}} & Basic & 35.6 (-2.1) & 43.9 (-2.0) & 71.5 (-3.9) & 57.8 (+2.2) \\
 & Smooth & 37.7 (-6.3) & 45.4 (-6.8) & 77.6 (-6.7) & 59.9 (-0.1) \\
 & Superpix & 39.3 (-9.7) & 47.7 (-12.3) & 76.9 (-4.0) & 61.6 (-0.2) \\ \hline
\multirow{3}{*}{\begin{tabular}[c]{@{}l@{}}Disc-Patch\end{tabular}} & Basic & 35.4 (-2.3) & 32.6 (-13.3) & 74.7 (-0.7) & 58.3 (+2.7) \\
 & Smooth & 38.6 (-5.4) & 45.8 (-6.4) & 78.8 (-5.5) & 61.7 (+1.7) \\
 & Superpix & 40.7 (-8.3) & 51.8 (-8.2) & 72.2 (-8.7) & 57.0 (-4.8) \\ \hline
\multirow{3}{*}{\begin{tabular}[c]{@{}l@{}}Disc-Crop\end{tabular}} & Basic & 45.1 (+7.4) & 54.0 (+8.1) & 76.5 (+1.1) & 55.5 (-0.1) \\
 & Smooth & 46.3 (+2.3) & 56.5 (+4.3) & 74.7 (-9.6) & 53.4 (-6.6) \\
 & Superpix & \textbf{50.6 (+1.6)} & 61.6 (+1.6) & 73.9 (-7.0) & 57.9 (-3.9) \\ \hline
\end{tabular}

\caption{Comparison of Random Crop, Discriminative Crop, Random Patch, and Discriminative Patch in terms of mIoU, FG-Precision, DR-/ NDR-Recall on VOC12. The difference between the aggregated and saliency performance is shown in parenthesis.}
\label{tab:cropping_quant}

\end{table}

\subsection{Can we do better than random cropping?}
Next, we explore different variations of random cropping and patching techniques that break the spatial structure of input images. Random patching is an erasure-based method similar to the idea of the cutout method \cite{devries2017improved}. The discriminative variations of random cropping (Disc-Crop)and patching (Disc-Patch) take the real values of CAM to complement the probability of selecting a crop or patch. See Appendix for details.
Table \ref{tab:cropping_quant} shows the results of these alternate methods. Random cropping and its discriminative variation (Disc-Crop) perform significantly better than the (basic) saliency method. However, the patch-based methods do not show comparative performance in terms of mIoU, FG-Precision, and DR-/NDR-Recall. One possible reason is that we used the original ``\textit{Model-org}'' classifier, which is not augmented with the patch-wise perturbations. Therefore, patching creates unnatural artifacts during inference, and the classifier fails to attribute the individual samples correctly. The discriminative versions of cropping and patching did not significantly outperform  the random versions.

\section{Related Works}

Current techniques for WS3 utilize CAMs as the foundation to produce segmentation maps. These methods can be broadly categorized into three types: (1) Modifying model architecture, (2) Iterative update-based methods, and (3) Modifying Loss functions. 

First, several methods that modify the model architecture for WS3 have been developed to overcome the well-known limitations of CAM \cite{kolesnikov2016seed, araslanov2020single, lee2021reducing} . For example, a global weighted rank (GWR) pooling layer was proposed in \cite{kolesnikov2016seed}  that neither underestimates the object size like global max pooling (GMP) nor overestimates it using GAP. Normalized global weighted pooling (nGWP) was also proposed in \cite{araslanov2020single} to replace the GAP layer, which helps to recover small segments, thus improving the mask precision. Another method FickleNet \cite{lee2019ficklenet} introduced stochastic aggregations in feature maps to produce the localization maps. However, changing the architecture can be difficult and restricts the types of models that are compatible with these methods.

The second set of methods aims to improve the seed performance of CAMs through iterative updates, such as erasure-based methods \cite{li2018tell, hou2018self, choe2020attention, wei2017object} and adversarial optimizations \cite{lee2021anti, wei2017object}. Specifically, erasure-based methods suggest erasing the most discriminative regions to unveil the non-discriminative regions, thus addressing some of the limitations of CAMs. On the other hand, AdvCAM \cite{lee2021anti} proposed an anti-adversarial optimization technique to exploit the boundary information with pixel-level affinity for capturing more regions of the target objects. One primary limitation of such methods is that the termination condition is not well-defined and often heuristically chosen. 

Finally, a third set of WS3 methods focus on modifying the loss function to improve the object coverage of CAMs. Specifically, the RIB \cite{lee2021reducing} demonstrates that an information bottleneck occurs in later layers as only the task-relevant information is passed to the output. As a result, CAMs which are computed at the last layer, have sparse coverage of the target object. A new loss function was proposed that encourages the transmission of information from non-discriminative regions for  classification, thus improving the quality of localization maps. 

Several prior works have utilized saliency maps for WS3, as documented in \cite{kolesnikov2016seed, shimoda2016distinct, sun2019saliency, zeng2019joint}. These studies primarily concentrate on enhancing segmentation map accuracy through post-processing techniques. However, their focus differs from our work on exploring the inherent potential of saliencies in overcoming the limitations associated with CAM-based approaches. Although these existing works contribute valuably to the field, they do not directly address the specific research questions that our study delves into – specifically, the comprehensive analysis of saliencies' effectiveness with respect to CAMs.

CAMs and Saliencies have also been extensively examined in the realm of explainability research that is focused on providing explanations of the model outputs, which can potentially satisfy regulatory experiments \cite{goodman2017european}, help practitioners debug their model \cite{casillas2013interpretability, cadamuro2016debugging} and identify unintended bias in the model \cite{lakkaraju2017interpretable, wang2015causal}.  Approaches based on activation maps fall under the CAM-based methods category \cite{zhou2016learning, selvaraju2016grad, chattopadhay2018grad, wang2020score}. Conversely, techniques relying on attribution maps belong to the saliency-like methods group \cite{simonyan2013deep, shrikumar2016not, springenberg2014striving, zeiler2014visualizing, smilkov2017smoothgrad, sundararajan2017axiomatic}.

\begin{table}[t]
\renewcommand{\arraystretch}{1.2}
\centering
\setlength\tabcolsep{3pt} 
\fontsize{9pt}{9}\selectfont
\begin{tabular}{cl|cccc}
\hline
\textbf{Method} & \multicolumn{1}{c|}{\textbf{B/G Resolve}} & \textbf{mIoU} & \textbf{FG-Prec} & \textbf{DR-Recall} & \textbf{NDR-Recall} \\ \hline
CAM & Basic & \textbf{28.82} & \textbf{41.16} & \textbf{83.59} & 31.46 \\ \hline
\multirow{2}{*}{Saliency} & Basic & 22.22 & 28.26 & 65.46 & 48.78 \\
 & Smooth & 25.46 & 31.94 & 73.02 & \textbf{52.65} \\ \hline
\multirow{2}{*}{{\begin{tabular}[c]{@{}c@{}}Random-\\ Crop\end{tabular}}} & Basic & 21.13 & 27.6 & 62.87 & 46.38 \\
 & Smooth & 26.58 & 33.83 & 72.09 & 52.22 \\ \hline
\end{tabular}

\caption{Quantitative comparison of CAM and Saliency on COCO dataset in terms of mIoU, Foreground Precision, and DR-/NDR-Recall.}
    \label{tab:cam_sal_coco}

\end{table}

\section{Discussion and Future Directions}

Table \ref{tab:cam_sal_coco} quantitatively evaluates the performance of competing methods on the MS COCO 2014 dataset. We compare the best-segmented map generated by each method by varying the global threshold across the range of $0.01$ to $0.50$. The segmented map with the highest mIoU value is selected for comparison. The Table illustrates that both saliency and random crop saliency outperform CAM in terms of NDR-Recall. This signifies that saliency-based approaches exhibit better recovery of the NDR region compared to CAM. However, CAM surpasses saliencies in terms of mIoU, FG-Precision, and DR-Recall. The smooth saliencies show comparable performance to CAM, which indicates the potential for improvement in the performance of saliencies by reducing its noisiness, especially when dealing with challenging datasets like the COCO dataset.

In conclusion, our paper proposes three novel evaluation metrics for WS3, namely NDR-Recall, DR-Recall, and FG-Precision, which can be used to assess the performance of alternative WS3 models in fixing the deficiencies of CAMs. We also revisit the potential of the use of saliency maps for WS3, which has been largely overlooked in the past, and demonstrate that simple post-processing steps, stochastic aggregation methods, and random cropping-based aggregation can significantly improve the quality of segmentation masks.

Although our work lays the foundation for future research in saliency maps for WS3, it's important to clarify that we are not the first to use saliencies for WS3, neither are we claiming state-of-the-art (SOTA) performance using stochastic aggregation methods when applied over saliencies. Instead, our focus is on presenting novel insights into the strengths and weakness of saliences w.r.t. CAMs from multiple perspectives, and showing how simple modifications to saliencies can effectively address the limitations inherent in CAMs. 

As newer techniques based on Vision Transformers \cite{xie2022clims, li2023transcam} and Foundation models such as Segment-Anything \cite{chen2023segment} are developed in the WS3 community to deliver SOTA performance, we anticipate future research to comprehensively understand their strengths and weaknesses building upon the metrics and analyses presented in our paper. Furthermore, while current post-processing methods in WS3 like CRF, PSA, and IRN are designed specifically to complement the limitations of CAM-based methods, we anticipate that researchers will build upon our findings to develop more advanced post-processing techniques for gradient-based WS3 methods. 

\bibliography{iclr2024_conference}
\bibliographystyle{iclr2024_conference}

\clearpage
\appendix
\appendix

\section{Comparison of CAMs and Saliency Maps Using Hyperplanes}

\subsection{Theoretical Proofs of CAM and SM-Hyperplanes}
The proof of the Remark \ref{remark:dr} and \ref{remark:hsr} are provided below.

\begin{remark}
If a  point $\mathbf{a} \in \mathcal{A}$ corresponding to a ground-truth pixel lies above $\mathcal{H}^c_{cam}$, i.e., $\mathbf{w}_{c}^{\text{T}} \mathbf{a}/Z - \tau_{cam} \geq 0$, then the pixel belongs to DR; otherwise, it belongs to NDR.
\end{remark}

\begin{proof}
The activation map $\mathbf{A}$ for an image $\mathbf{I}$ can be sampled at any arbitrary ground-truth pixel $(i, j) \in \mathcal{S}^c_{GT}$ such that $\mathbf{A}_{(i,j)} \in \mathcal{A}$. Therefore, the CAM score for the $c$-th Class at pixel $(i, j)$ can be computed as: $\text{CAM}_c(i, j) = \mathbf{w}_{c}^{\text{T}}\mathbf{A}_{(i,j)}/Z$.

Now, if the pixel belongs to the discriminative region (DR), then by definition \ref{def:dr_ndr} we get the following:
\begin{align}
    &CAM_c(i, j) = \frac{\mathbf{w}_{c}^{\text{T}}}{Z}\mathbf{A}_{(i,j)} \geq \tau_{cam} \nonumber\\
    &\implies \frac{\mathbf{w}_{c}^{\text{T}}}{Z}\mathbf{A}_{(i,j)} - \tau_{cam} \geq 0
\end{align}
This by definition of $\mathcal{H}^c_{cam}$ (see Definition \ref{def:cam_hyperplane}) suggests that the activation value $\mathbf{A}_{(i,j)} \in \mathcal{A}$ lies above $\mathcal{H}^c_{cam}$.

Conversely, if the ground-truth pixel belongs to the non-discriminative region (NDR), then by definition \ref{def:dr_ndr} we get the following:

\begin{align}
    &CAM_c(i, j) = \frac{\mathbf{w}_{c}^{\text{T}}}{Z}\mathbf{A}_{(i,j)} < \tau_{cam} \nonumber\\
    &\implies \frac{\mathbf{w}_{c}^{\text{T}}}{Z}\mathbf{A}_{(i,j)} - \tau_{cam} < 0
\end{align}

Similarly, from the definition of $\mathcal{H}^c_{cam}$ (see Definition \ref{def:cam_hyperplane}), the activation value $\mathbf{A}_{(i,j)} \in \mathcal{A}$ lies below $\mathcal{H}^c_{cam}$.

Therefore, we can say in general if an arbitrary point $\mathbf{a}$ corresponding to a ground-truth pixel lies above $\mathcal{H}^c_{cam}$, it belongs to the discriminative region (DR); otherwise it belongs to NDR.
\end{proof}

\begin{remark}
If a point $\mathbf{a'}$ corresponding to a ground-truth pixel lies on the outer sides of $\mathcal{H}^c_{sm}$, i.e., $|\mathbf{w}_{c}^{\text{T}} \mathbf{a'}| - \tau_{sm} \geq 0$, then the point belongs to HSR; otherwise, it belongs to LSR.
\end{remark}

\begin{proof}
Similar to CAM, the gradient of the GAP w.r.t. the image $\frac{\partial \text{GAP}(\mathbf{A})}{\partial \mathbf{I}}$ can be sampled at any arbitrary ground-truth pixel $(i, j) \in \mathcal{S}^c_{GT}$ such that $\frac{\partial \text{GAP}(\mathbf{A})}{\partial \mathbf{I}}\big|_{(i, j)} \in \mathcal{A'}$. Therefore, the Saliency map score for the $c$-th Class at pixel $(i, j)$ can be computed as: 
\begin{equation}
\text{SM}_c(i, j) = \Bigg|\mathbf{w}_{c}^{\text{T}}\frac{\partial \text{GAP}(\mathbf{A})}{\partial \mathbf{I}}\big|_{(i, j)} \Bigg|
\end{equation}

Now, if the pixel belongs to the high saliency region (HSR), then by definition \ref{def:hsr} we get the following:
\begin{align}
    &SM_c(i, j) = \Bigg|\mathbf{w}_{c}^{\text{T}}\frac{\partial \text{GAP}(\mathbf{A})}{\partial \mathbf{I}}\big|_{(i, j)}\Bigg| \geq \tau_{sm} \nonumber\\
    &\implies \Bigg|\mathbf{w}_{c}^{\text{T}}\frac{\partial \text{GAP}(\mathbf{A})}{\partial \mathbf{I}}\big|_{(i, j)}\Bigg| - \tau_{sm} \geq 0 
\end{align}

This by definition of the $\mathcal{H}^c_{sm}$ (see Definition \ref{def:sm_hyperplane}) suggests that the gradient $\frac{\partial \text{GAP}(\mathbf{A})}{\partial \mathbf{I}}\big|_{(i, j)} \in \mathcal{A'}$ lies on the outer sides of the $\mathcal{H}^c_{sm}$.

Similarly, if the pixel belongs to the low saliency region (LSR), then by definition \ref{def:hsr} we get the following:
\begin{align}
    &SM_c(i, j) = \Bigg|\mathbf{w}_{c}^{\text{T}}\frac{\partial \text{GAP}(\mathbf{A})}{\partial \mathbf{I}}\big|_{(i, j)}\Bigg| < \tau_{sm} \nonumber\\
    &\implies \Bigg|\mathbf{w}_{c}^{\text{T}}\frac{\partial \text{GAP}(\mathbf{A})}{\partial \mathbf{I}}\big|_{(i, j)}\Bigg| - \tau_{sm} < 0 
\end{align}

This by definition of $\mathcal{H}^c_{sm}$ (see Definition \ref{def:sm_hyperplane}) suggests that the gradient $\frac{\partial \text{GAP}(\mathbf{A})}{\partial \mathbf{I}}\big|_{(i, j)} \in \mathcal{A'}$ lies on the inner sides of the $\mathcal{H}^c_{sm}$.

Therefore, we can say in general if an arbitrary point $\mathbf{a'}$ corresponding to a ground-truth pixel lies on the outer sides the $\mathcal{H}^c_{sm}$, it belongs to the high-saliency region (HSR); otherwise it belongs to low saliency region (LSR).
\end{proof}


\subsection{Visual Comparison for more representative images from VOC}
\begin{figure*}[t]
\centering
\begin{subfigure}{0.17\linewidth}
   \includegraphics[width=\linewidth]{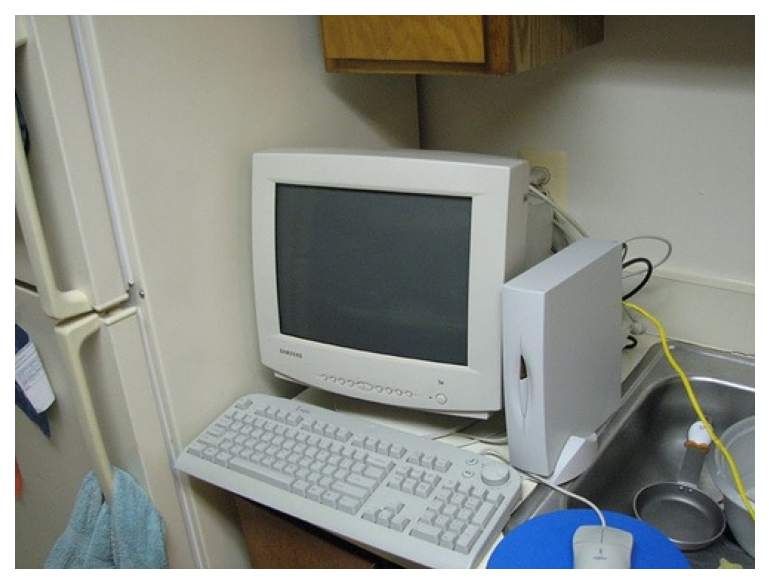}
\end{subfigure}
\begin{subfigure}{0.17\linewidth}
   \includegraphics[width=\linewidth]{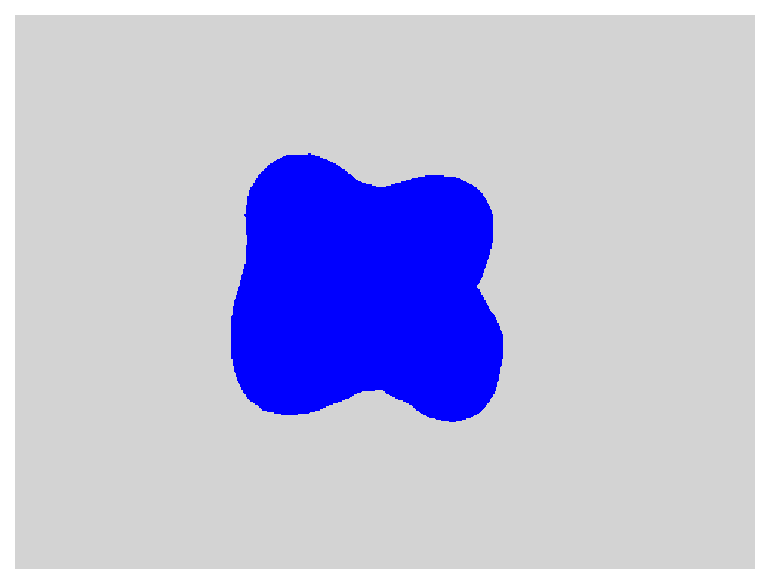}
\end{subfigure}
\begin{subfigure}{0.17\linewidth}
   \includegraphics[width=\linewidth]{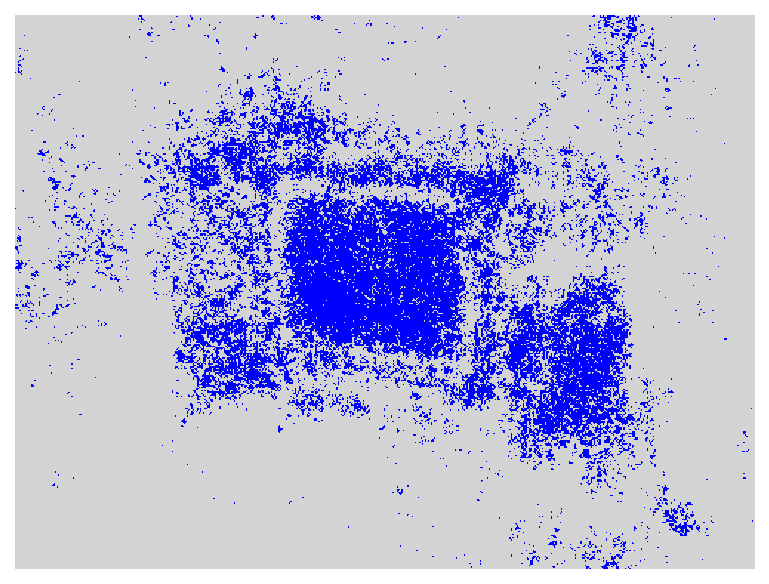}
\end{subfigure}
\begin{subfigure}{0.22\linewidth}
   \includegraphics[width=\linewidth]{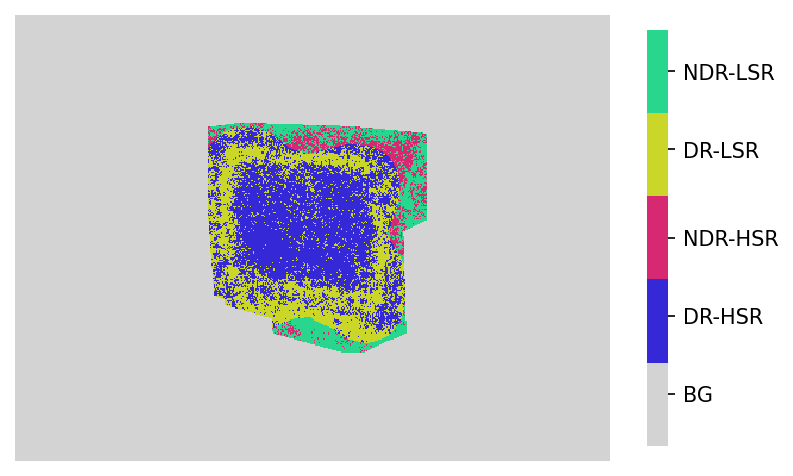}
\end{subfigure}
\begin{subfigure}{0.2\linewidth}
   \includegraphics[width=\linewidth]{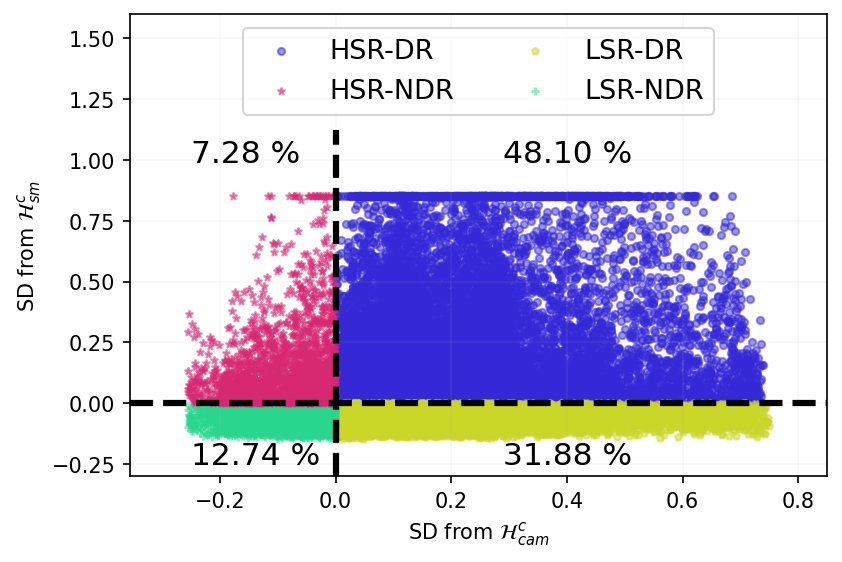}
\end{subfigure}

\begin{subfigure}{0.17\linewidth}
   \includegraphics[width=\linewidth]{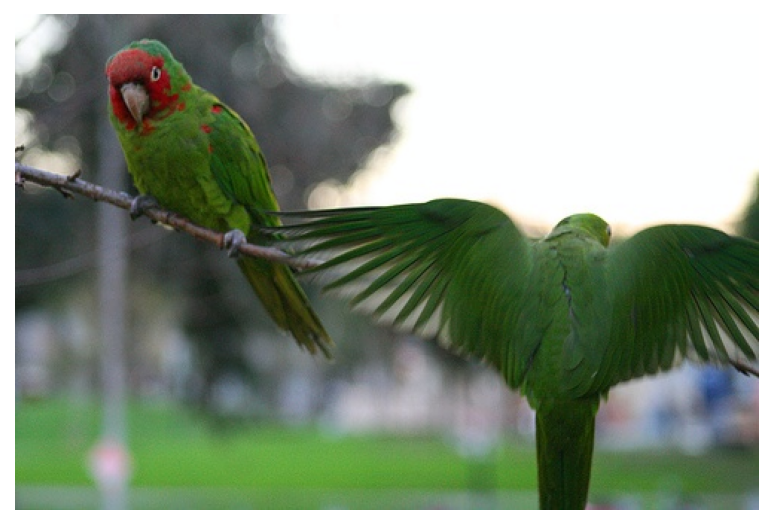}
\end{subfigure}
\begin{subfigure}{0.17\linewidth}
   \includegraphics[width=\linewidth]{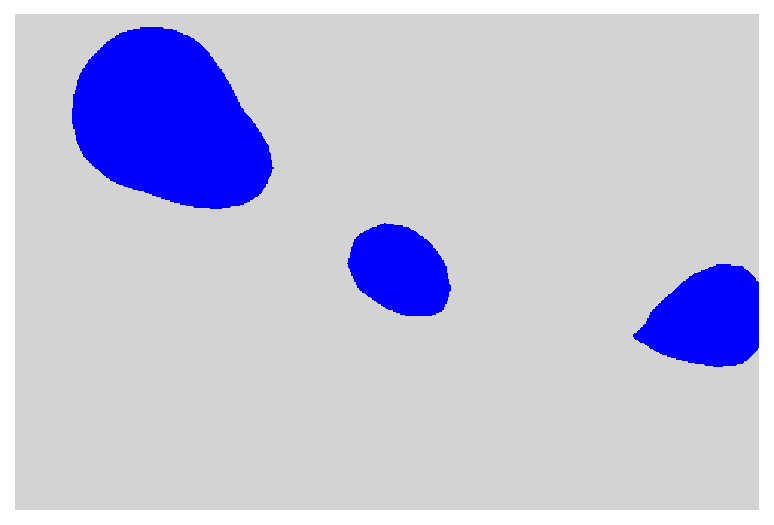}
\end{subfigure}
\begin{subfigure}{0.17\linewidth}
   \includegraphics[width=\linewidth]{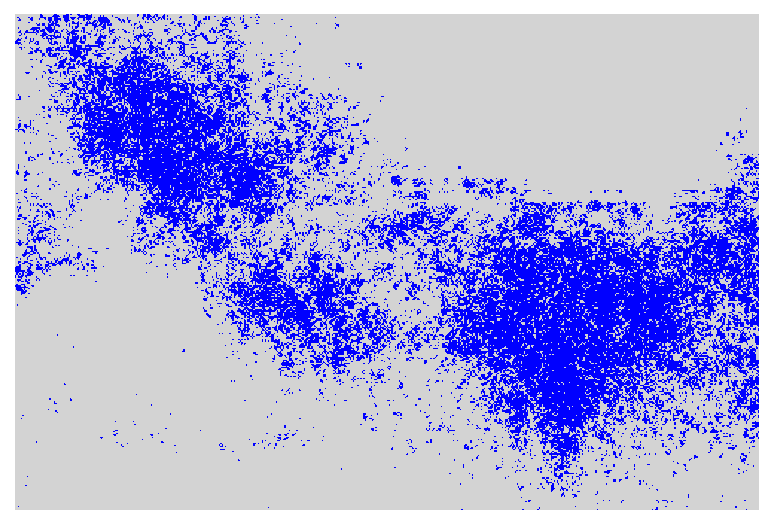}
\end{subfigure}
\begin{subfigure}{0.22\linewidth}
   \includegraphics[width=\linewidth]{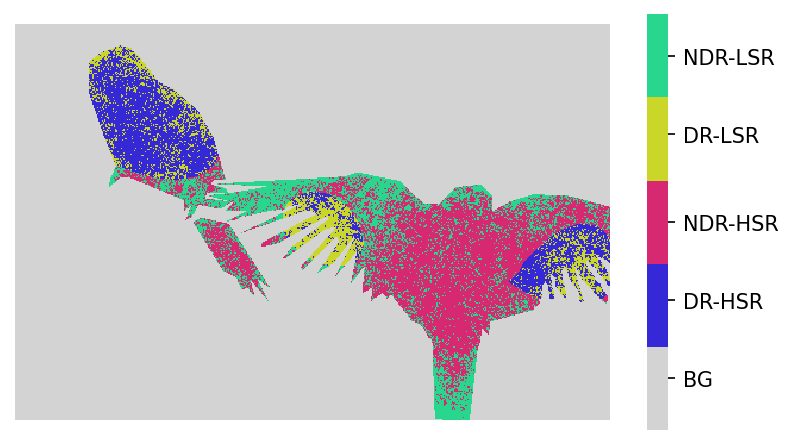}
\end{subfigure}
\begin{subfigure}{0.2\linewidth}
   \includegraphics[width=\linewidth]{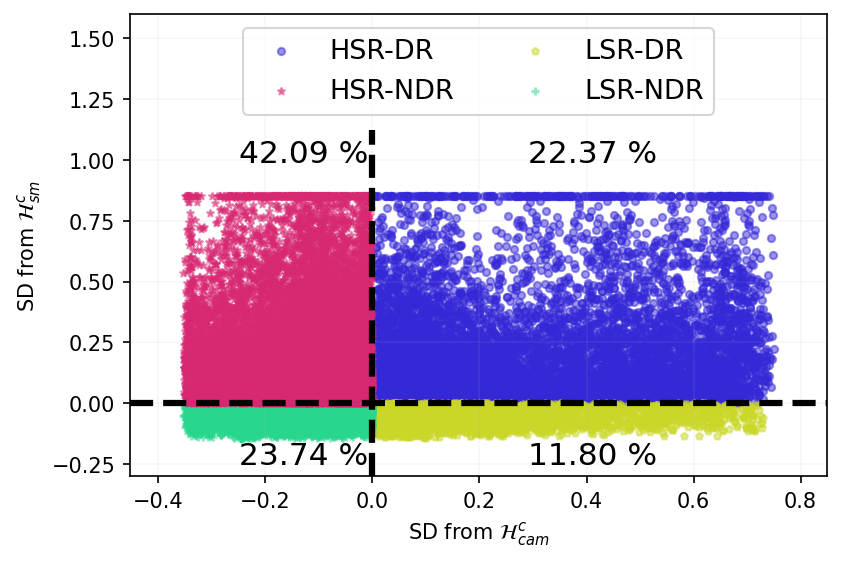}
\end{subfigure}

\begin{subfigure}{0.17\linewidth}
   \includegraphics[width=\linewidth, height=80px]{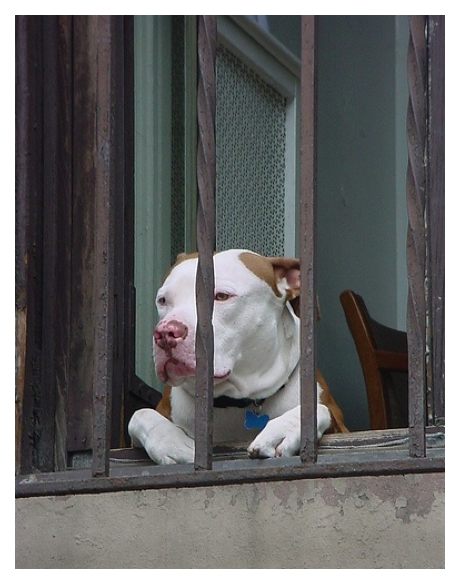}
\end{subfigure}
\begin{subfigure}{0.17\linewidth}
   \includegraphics[width=\linewidth, height=80px]{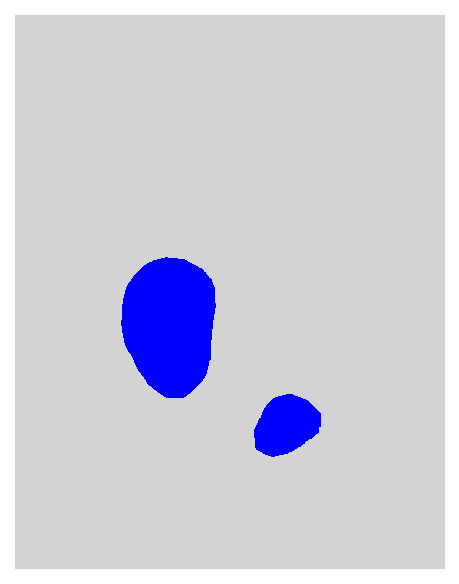}
\end{subfigure}
\begin{subfigure}{0.17\linewidth}
   \includegraphics[width=\linewidth, height=80px]{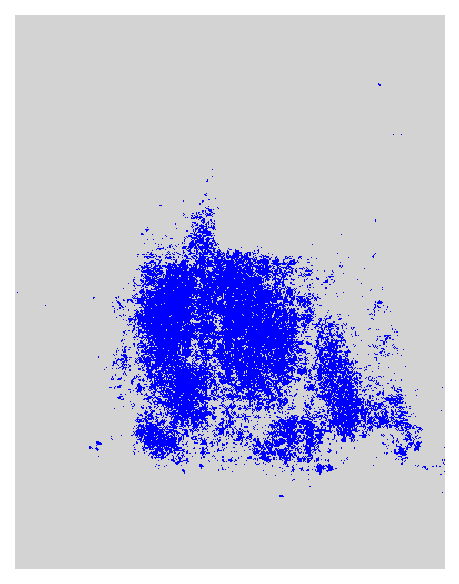}
\end{subfigure}
\begin{subfigure}{0.22\linewidth}
   \includegraphics[width=\linewidth, height=80px]{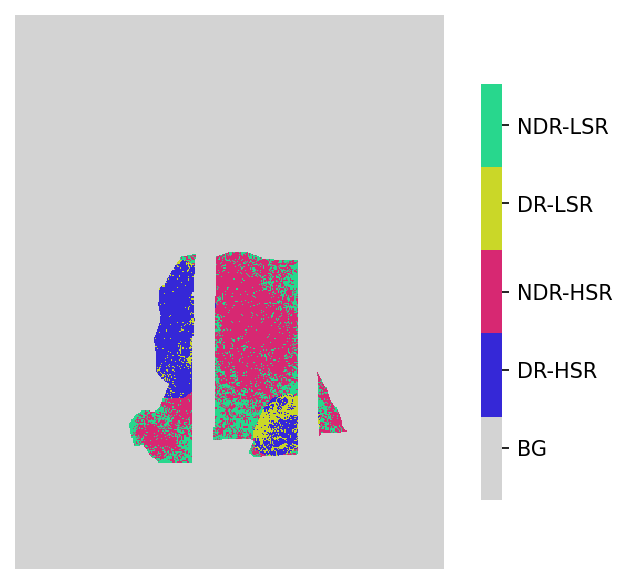}
\end{subfigure}
\begin{subfigure}{0.2\linewidth}
   \includegraphics[width=\linewidth]{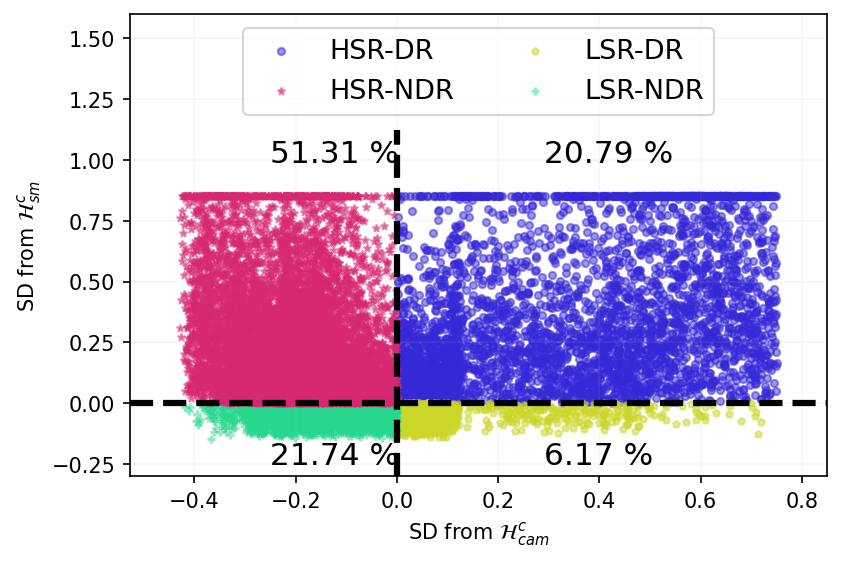}
\end{subfigure}

\begin{subfigure}{0.17\linewidth}
   \includegraphics[width=\linewidth]{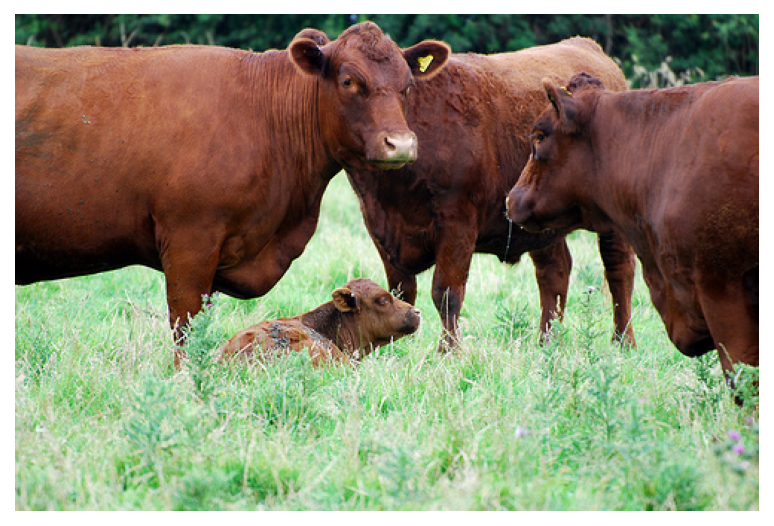}
\end{subfigure}
\begin{subfigure}{0.17\linewidth}
   \includegraphics[width=\linewidth]{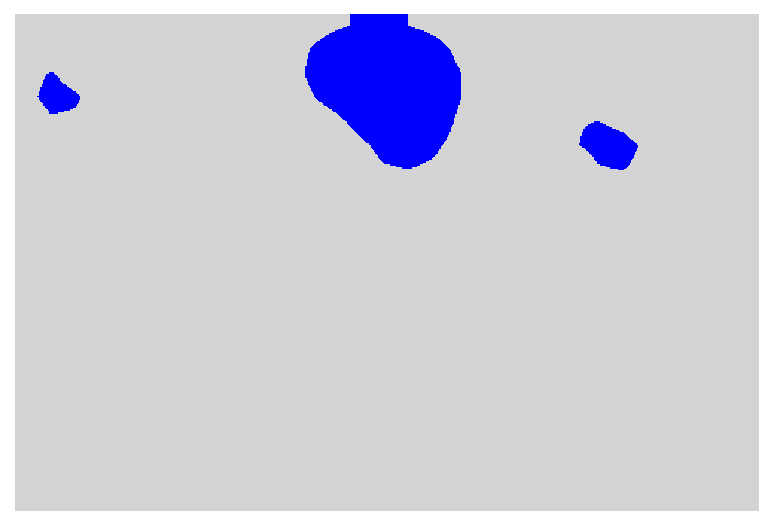}
\end{subfigure}
\begin{subfigure}{0.17\linewidth}
   \includegraphics[width=\linewidth]{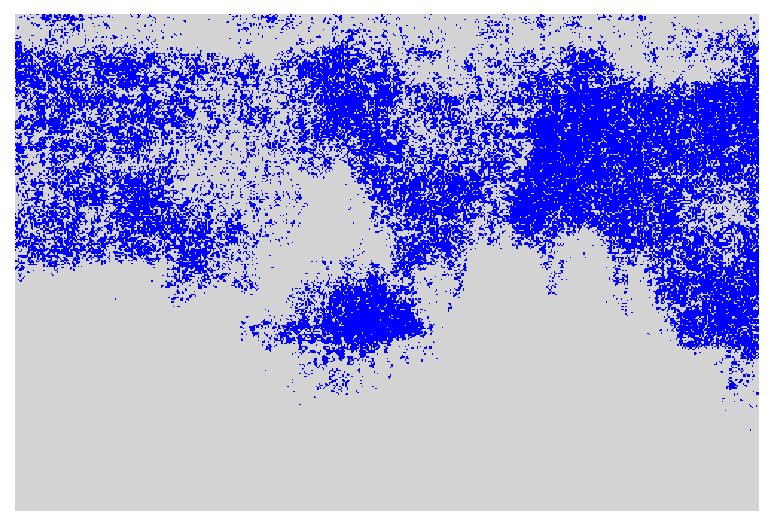}
\end{subfigure}
\begin{subfigure}{0.22\linewidth}
   \includegraphics[width=\linewidth]{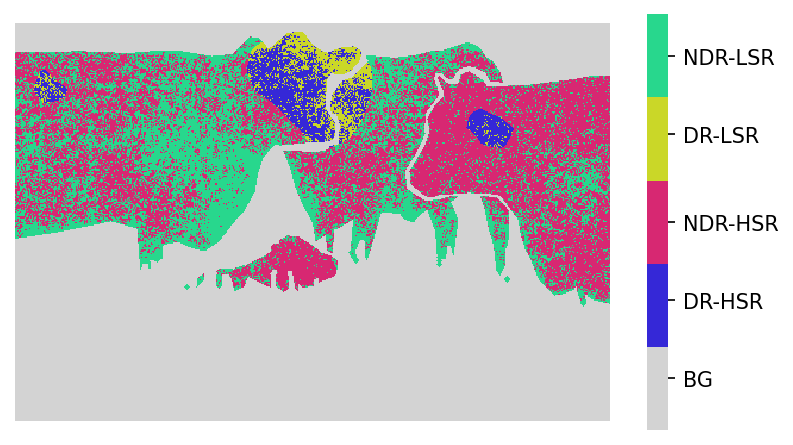}
\end{subfigure}
\begin{subfigure}{0.2\linewidth}
   \includegraphics[width=\linewidth]{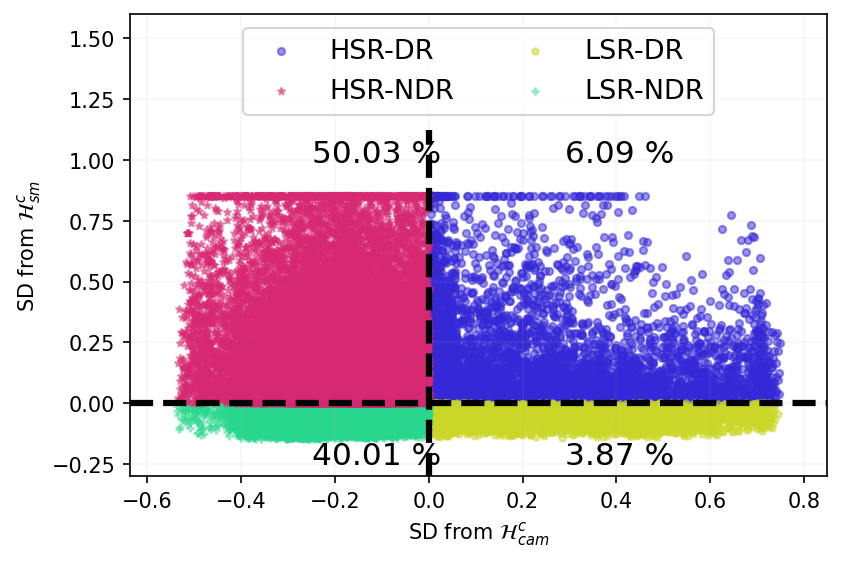}
\end{subfigure}

\begin{subfigure}{0.17\linewidth}
   \includegraphics[width=\linewidth]{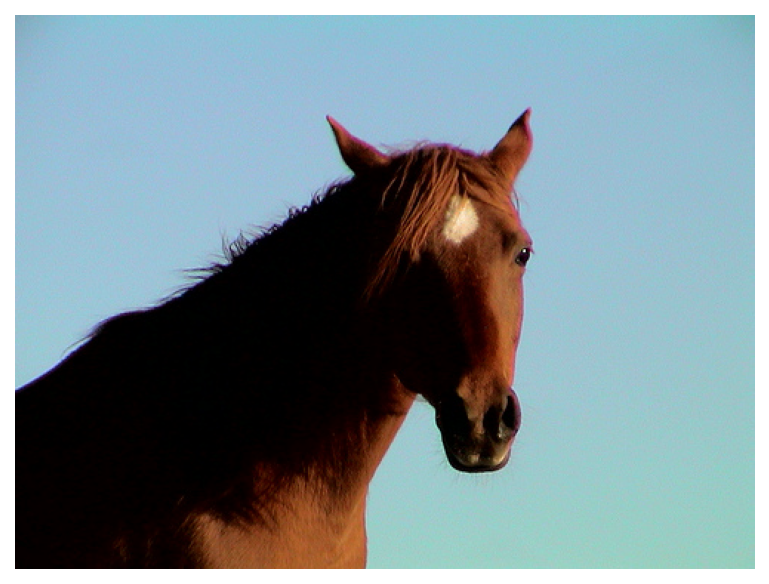}
\end{subfigure}
\begin{subfigure}{0.17\linewidth}
   \includegraphics[width=\linewidth]{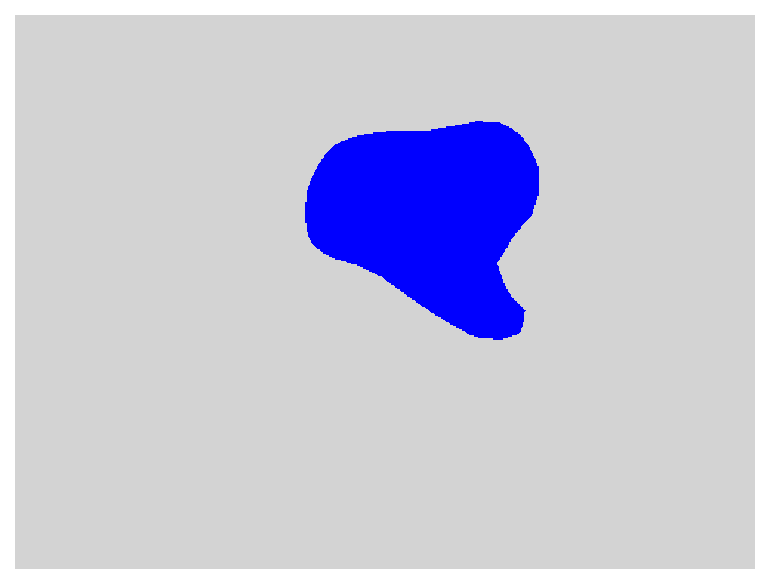}
\end{subfigure}
\begin{subfigure}{0.17\linewidth}
   \includegraphics[width=\linewidth]{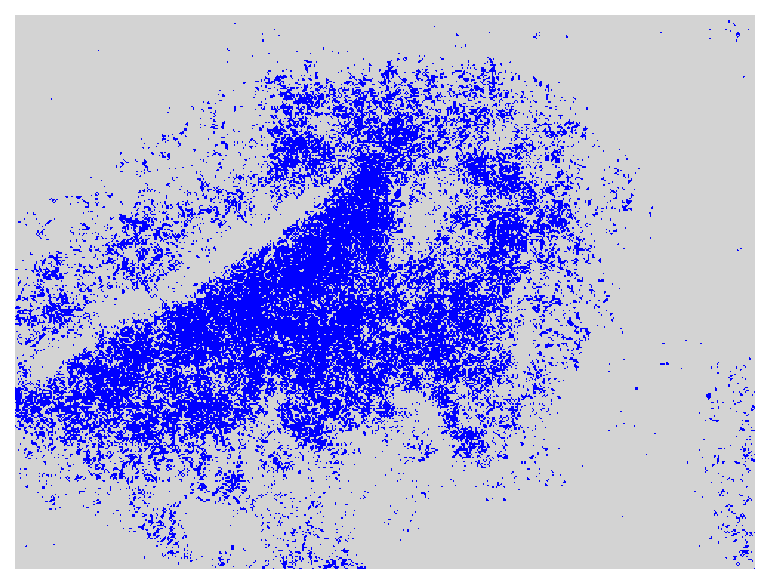}
\end{subfigure}
\begin{subfigure}{0.22\linewidth}
   \includegraphics[width=\linewidth]{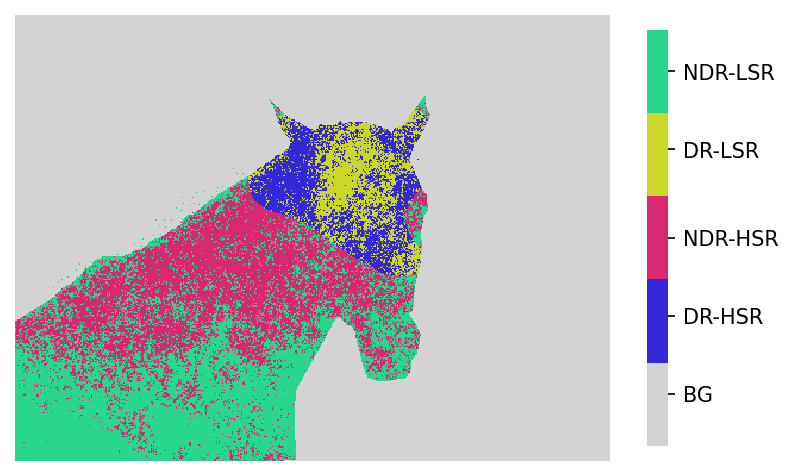}
\end{subfigure}
\begin{subfigure}{0.2\linewidth}
   \includegraphics[width=\linewidth]{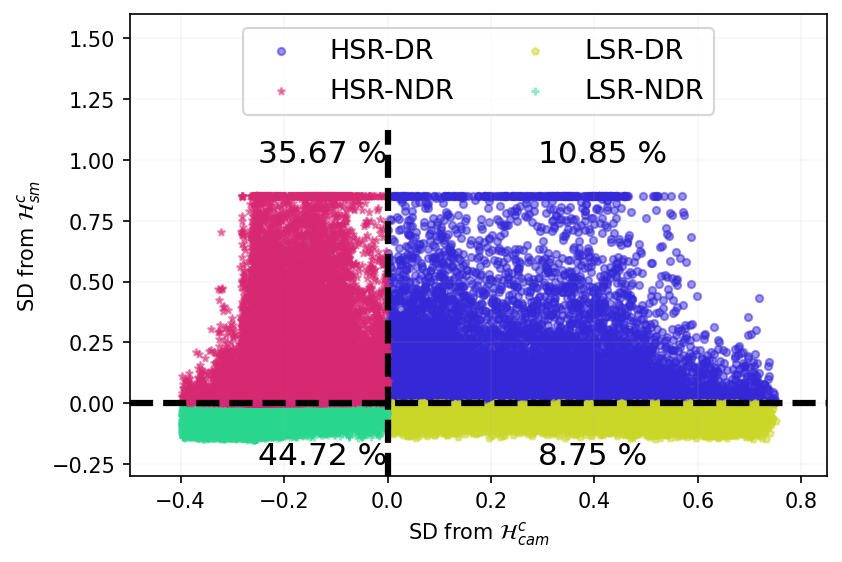}
\end{subfigure}

\begin{subfigure}{0.17\linewidth}
   \includegraphics[width=\linewidth]{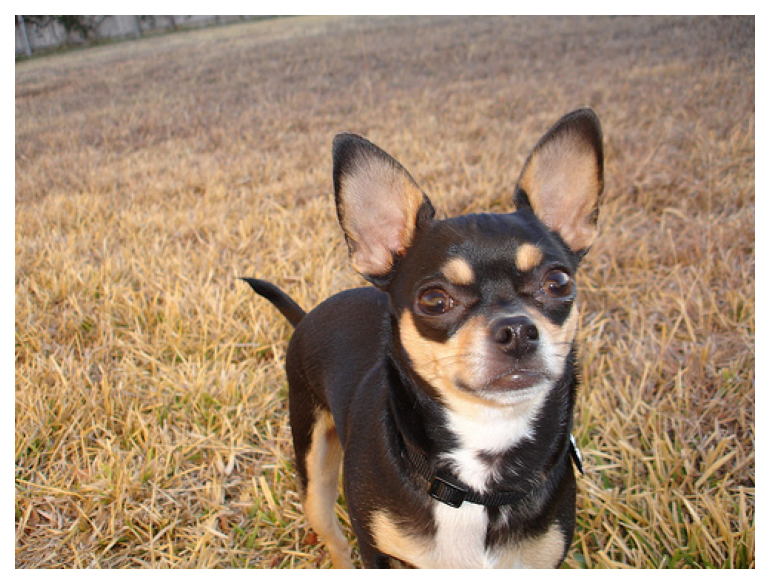}
\end{subfigure}
\begin{subfigure}{0.17\linewidth}
   \includegraphics[width=\linewidth]{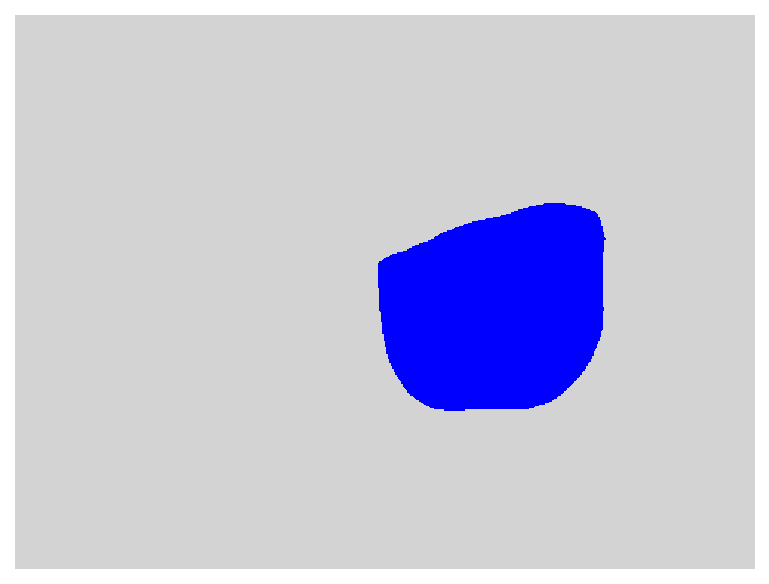}
\end{subfigure}
\begin{subfigure}{0.17\linewidth}
   \includegraphics[width=\linewidth]{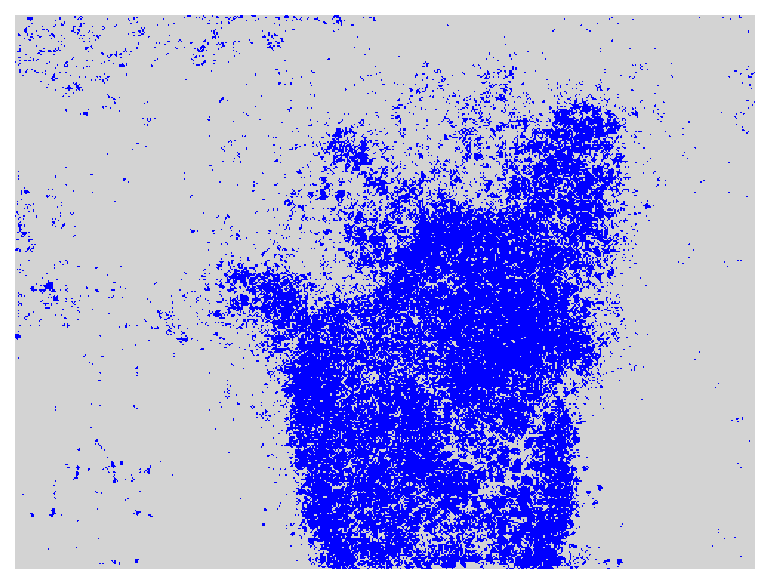}
\end{subfigure}
\begin{subfigure}{0.22\linewidth}
   \includegraphics[width=\linewidth]{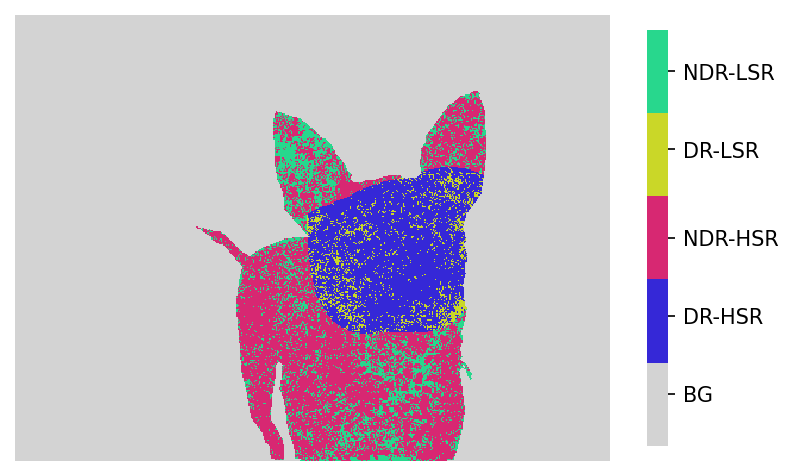}
\end{subfigure}
\begin{subfigure}{0.2\linewidth}
   \includegraphics[width=\linewidth]{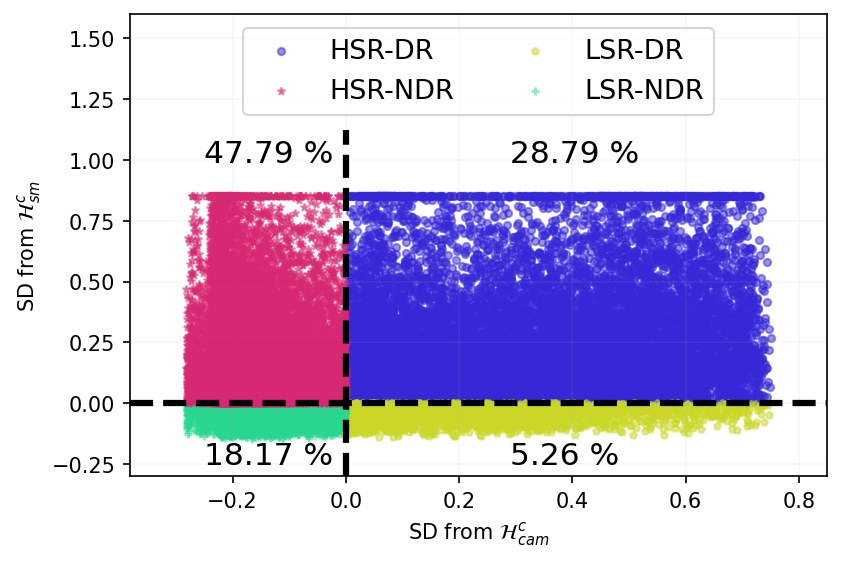}
\end{subfigure}

\begin{subfigure}{0.17\linewidth}
   \includegraphics[width=\linewidth]{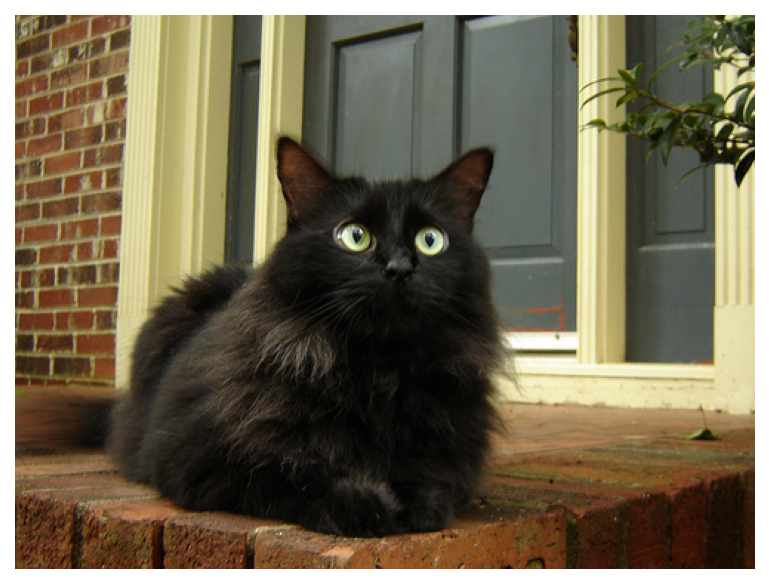}
   \caption{Original Image}
\end{subfigure}
\begin{subfigure}{0.17\linewidth}
   \includegraphics[width=\linewidth]{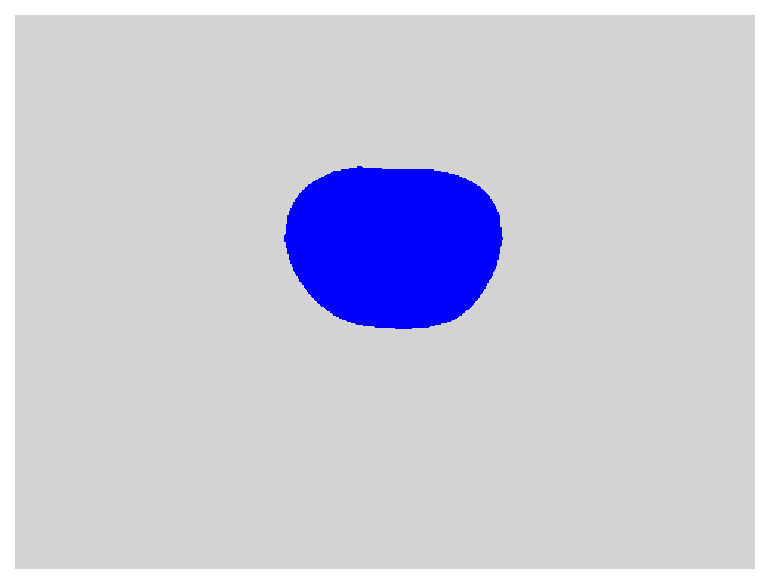}
   \caption{CAM}
\end{subfigure}
\begin{subfigure}{0.17\linewidth}
   \includegraphics[width=\linewidth]{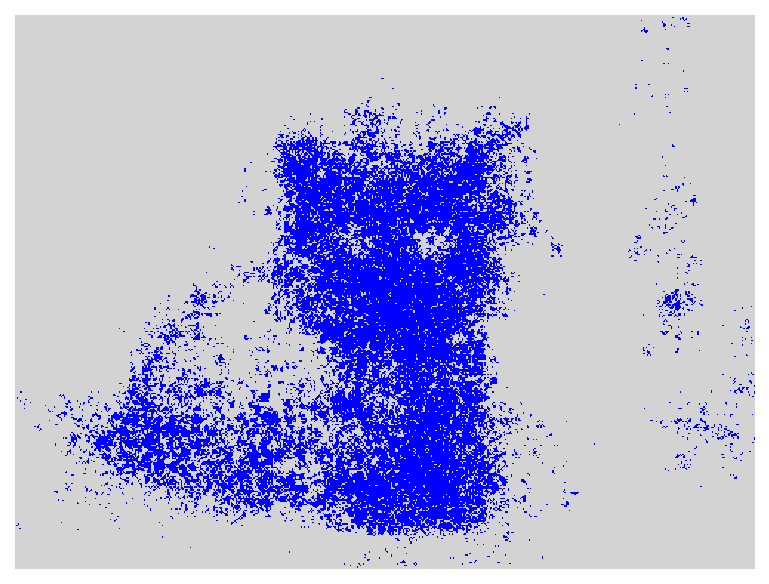}
   \caption{Saliency Map (SM)}
\end{subfigure}
\begin{subfigure}{0.22\linewidth}
   \includegraphics[width=\linewidth]{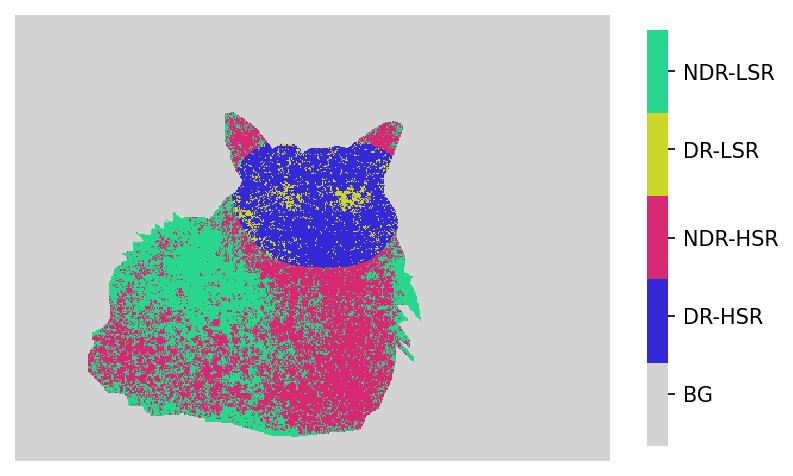}
   \caption{CAM vs. SM - Image}
\end{subfigure}
\begin{subfigure}{0.2\linewidth}
   \includegraphics[width=\linewidth]{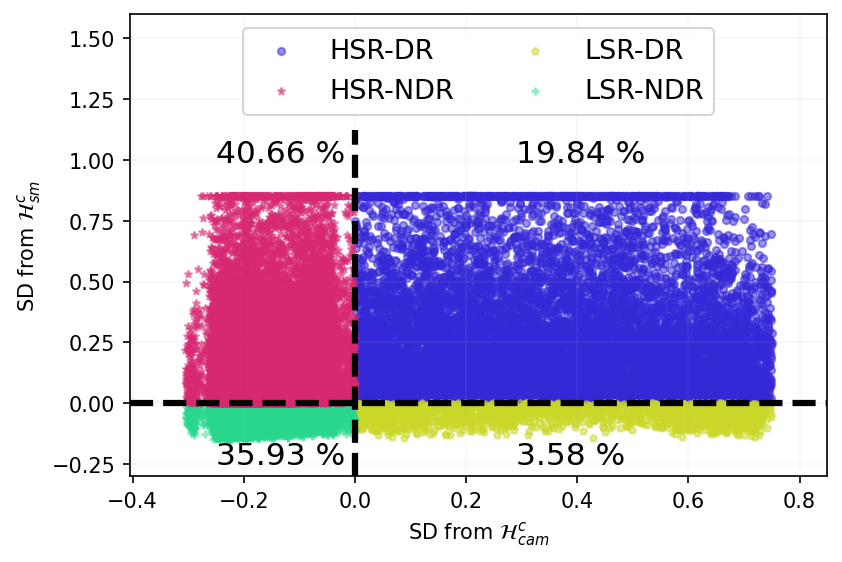}
   \caption{CAM vs. SM - Distances}
\end{subfigure}
\caption{A visual comparison of CAMs and saliency maps (SMs) for more representative images from the VOC12 dataset.}
\label{fig:viz_hyperplane}
\end{figure*}

Figure \ref{fig:viz_hyperplane} presents a visual comparison of CAMs and Saliencies using the hyperplanes on more representative images from the PASCAL VOC 12 dataset (similar to Figure \ref{fig:hyperplane} from the main paper). For this experiment, we choose the value of $\tau_{cam} = 0.25$ and $\tau_{sm} = 0.15$.

\section{Experimental Details}

\subsection{Dataset Description}
We compared different competing approaches quantitatively and qualitatively by conducting experiments on MNIST, PASCAL VOC '12, and MS COCO '14 datasets. 
\subsubsection{MNIST Segmentation Dataset:}
We generate the ground-truth segmentation masks by filtering the non-zero pixels of the MNIST images. For our experiments, we used an upsampled version of the original MNIST dataset, where we used ``nearest neighbor" interpolation to upsample the dataset to $128 \times 128$ dimension. Furthermore, we used 60,000 training set and 10,000 test set images with segmentation masks for our experiments in Section 5.

\subsubsection{PASCAL VOC '12 Dataset:}
The PASCAL VOC 2012 dataset contains 10,582 training images, 1,449 validation images, and 1,456 test images with objects from 20 classes. We compared the methods by evaluating the performance of the 1,464 segmented images using the approach adopted in recent WS3 research.

\subsubsection{MS COCO '14 Dataset:}
The MS COCO 2014 dataset contains 82,783 training and 40,504 validation images with objects from 80 classes. We evaluated the competing approaches on approximately 82K training images from the MS COCO 2014 dataset.

\subsection{Model Description}
We fine-tuned a classification network to accurately extract segmented seeds, utilizing ResNet50 as the backbone network, which is pre-trained on ImageNet. In order to maintain consistency with prior research, we incorporated various augmentations during the fine-tuning process, such as resizing to (320, 640), applying a horizontal flip with a 0.5 probability, and cropping with a maximum size of 512. We developed and fine-tuned three separate classification models to explore the impact of different perturbations during the fine-tuning stage. \textit{Model-org} model is fine-tuned only with the aforementioned augmentations.
During fine-tuning, we perturb the input image with binary noise to create additional augmentations for \textit{Model-pert-binary} classification model. Formally, 
\begin{align*}
    &\tilde{\mathbf{I}} = \mathbf{I} \odot m \nonumber \\
    &m \sim \text{Bernoulli}(p), \quad \text{where, }p=0.9
\end{align*}
$\tilde{\mathbf{I}}$ is the training image that is perturbed with the binary mask $m$. The mask has a binary probability $p=0.9$ to set each pixel. Similar to Model-pert-binary, we additionally perturb the input image with  Gaussian noise for the \textit{Model-pert-gaussian} classification model. Formally,
\begin{align*}
    &\tilde{\mathbf{I}} = \mathbf{I} + \epsilon \nonumber \\
    &\epsilon \sim \mathcal{N}(0, \sigma), \quad \text{where, }\sigma=0.15
\end{align*}
$\tilde{\mathbf{I}}$ is the training image that is perturbed with the Gaussian noise $\epsilon$. The noise level (perturbation) is controlled by the standard deviation $\sigma = 0.15$.

\subsection{Background Resolve Techniques}
\subsubsection{Basic Background Resolve}
This is the most common approach in recent research that uses a simple strategy for distinguishing between foreground and background classes. This is done by setting a global threshold that discerns the background class and then assigning classes based on the highest real values among the foreground classes.


\subsubsection{Kernel Smoothing}
The technique of Kernel Smoothing has been utilized to smooth the gradients of the vanilla saliency maps by applying a Gaussian kernel with a kernel size of 13 and a standard deviation of 5. Following this, a global threshold has been selected to distinguish foreground classes from the background. This has been achieved by considering the maximum real values of the smoothed saliencies for the target classes. This approach has been adopted to enhance the accuracy of the saliency maps by smoothing the gradients as a post-processing step.

\subsubsection{Superpixel-based Background Resolve}
Superpixels consist of clusters of pixels that exhibit similar characteristics. In contrast to the conventional method of assigning a label to each individual pixel, superpixel-based smoothing allocates a label to each superpixel, effectively reducing the noise and scatteredness present in saliency maps. We employed Felzenszwalb's efficient graph-based superpixel algorithm \cite{felzenszwalb2004efficient} to compute the superpixels. To designate a class label for each superpixel, we initially calculated the mean saliencies for every superpixel. And then, using a global threshold of $0.3$, we determine whether a superpixel is part of the foreground or background. We assigned target classes for foreground superpixels based on the highest mean gradients concerning the target classification score.

\begin{figure*}[h]
\begin{center}
   \includegraphics[width=1.00\linewidth]{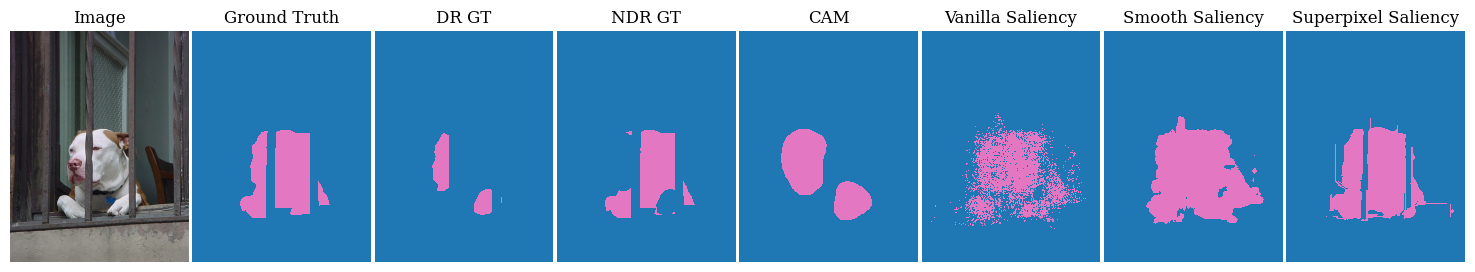}
   \includegraphics[width=1.00\linewidth]{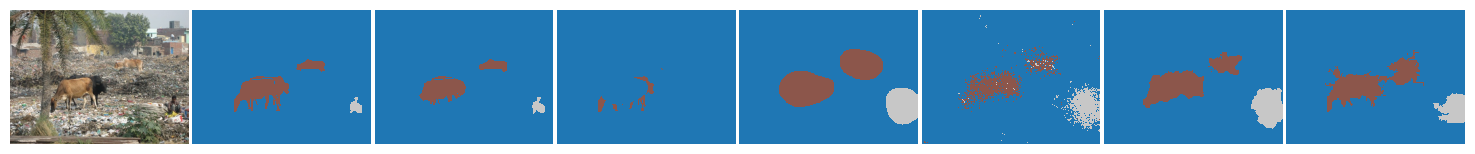}
   \includegraphics[width=1.00\linewidth]{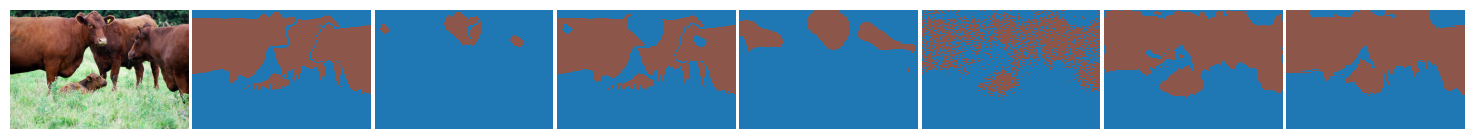}
   \includegraphics[width=1.00\linewidth]{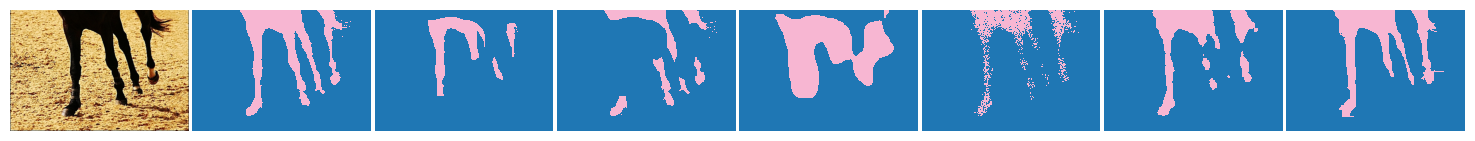}
   \includegraphics[width=1.00\linewidth]{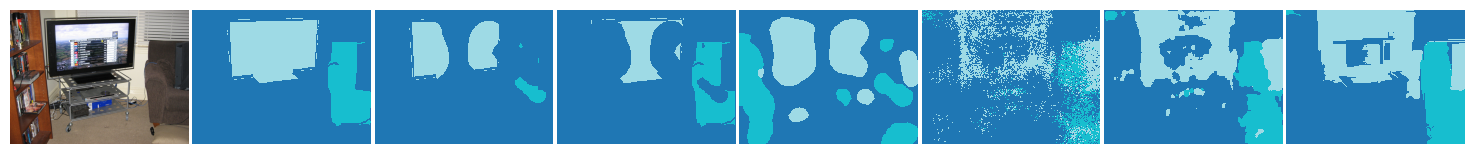}
   \includegraphics[width=1.00\linewidth]{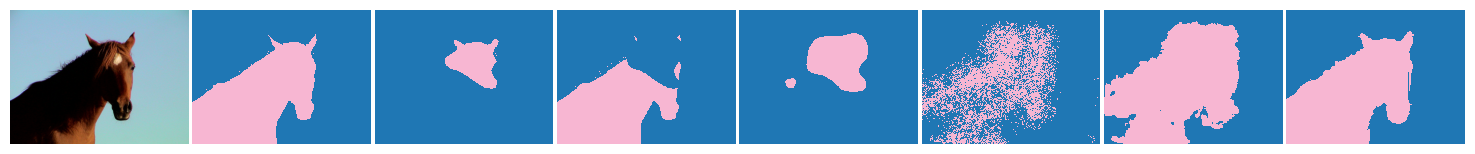}
   \includegraphics[width=1.00\linewidth]{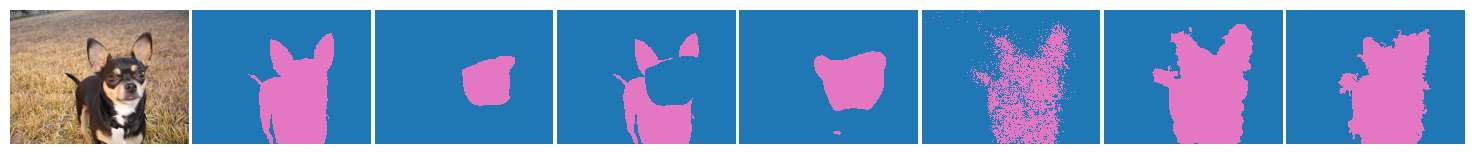}
   \includegraphics[width=1.00\linewidth]{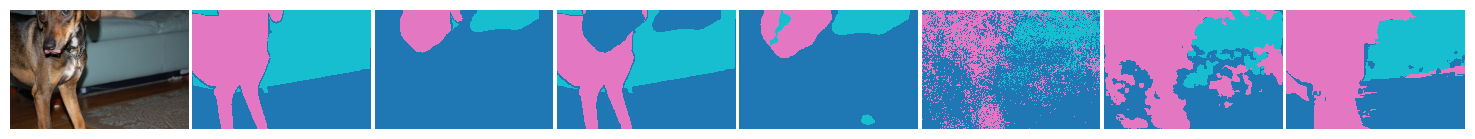}
   \includegraphics[width=1.00\linewidth]{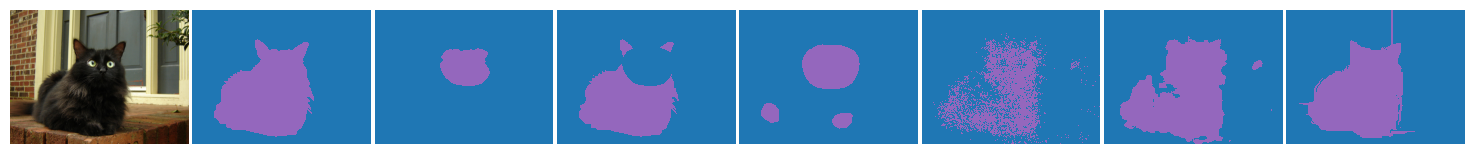}
   \includegraphics[width=1.00\linewidth]{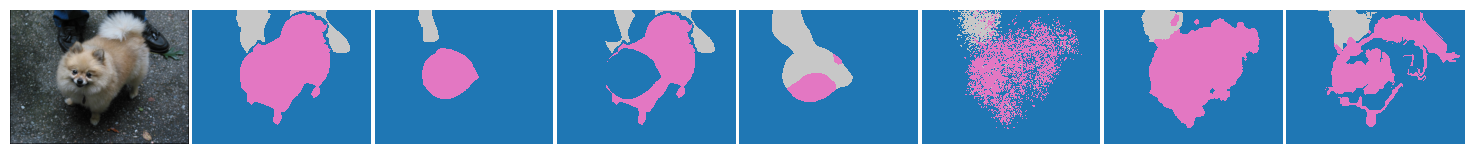}
   \includegraphics[width=1.00\linewidth]{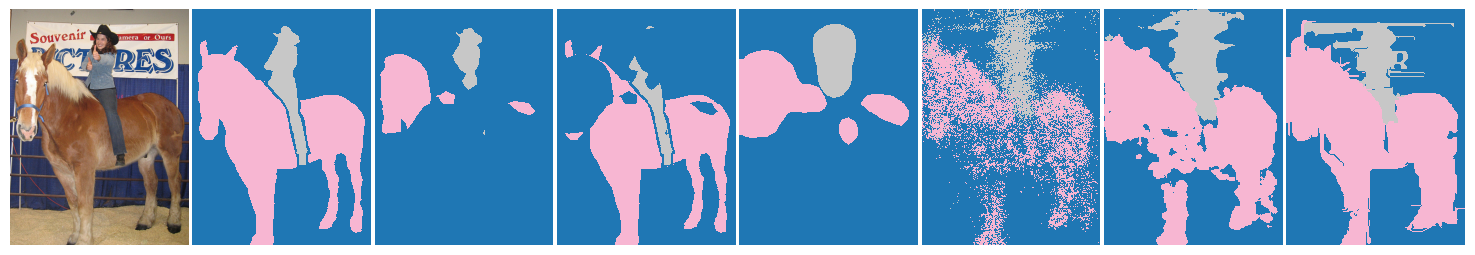}
\end{center}
   \caption{Visual comparison between CAM and Vanilla Saliency with different background resolves.}
\label{fig:supple_viz_compare_cam_sal}
\end{figure*}

In an effort to better understand the performance of background resolution techniques, Figure \ref{fig:supple_viz_compare_cam_sal} presents a visual comparison between CAM and Vanilla Saliency with different resolution methods, namely Basic, Smooth and Superpixel. The basic background resolution is represented by ``Vanilla Saliency". For this experiment, we set a global threshold of 0.15 to differentiate the foreground from the background. Building upon the insights gathered from Section \ref{sec:saliency_post_proc}, we observed the following implications for each background resolution approach. Employing vanilla saliency with Basic background resolution results in noisy and scattered saliency maps, demonstrating its limitations in providing clear object segmentation. Utilizing Kernel smoothing generates smooth saliencies, which offers an improvement over the Basic technique. However, this approach still struggles with unclear object boundaries, making it difficult to precisely locate objects within the image (First and Sixth row of Figure \ref{fig:supple_viz_compare_cam_sal}). The Superpixel-based background resolution effectively smooths the saliencies while maintaining clear and distinguishable object boundaries, presenting a more refined solution (First, Fourth, and Sixth row of Figure \ref{fig:supple_viz_compare_cam_sal}). Nonetheless, this method has its drawbacks, as the resulting saliencies heavily rely on the superpixel shapes and the algorithm’s ability to identify them accurately. Consequently, any slight deviation from the correct superpixel shape can cause this background resolution technique to fail in capturing the entire body of the target object (Eight, tenth, and eleventh row of Figure \ref{fig:supple_viz_compare_cam_sal}).

\section{Stochastic Aggregation for Saliencies}
\subsection{SmoothGrad and BinaryMask}
To reduce noise, \cite{smilkov2017smoothgrad} proposes a stochastic aggregation-based saliency map, namely SmoothGrad, where Gaussian noise is added to the input image to construct a neighborhood of the input image. Then, $n$ different random samples are selected from the neighborhood, and the saliencies of all the samples are averaged to generate the final saliency, which is much smoother than the Vanilla Saliency.

In this paper, we explored another variation of input noise perturbation, namely BinaryMask, where, instead of adding Gaussian noise to the input image, we multiply the image by a binary noise. We can formally define both these methods as follows:
\begin{align}
    &\tilde{\text{SM}}_c(x) = \frac{1}{n} \sum_{1}^{n} \text{SM}_c(\tilde{\mathbf{I}}) \label{eq:stochastic_agg}\\
    &\tilde{\mathbf{I}} = \mathbf{I} + \epsilon; \:\: \epsilon \sim \mathcal{N}(0, \sigma^2), \text{for SmoothGrad} \\
    &\tilde{\mathbf{I}} = \mathbf{I} \odot m; \:\: m \sim \text{Bernoulli}(p), \text{for BinaryMask}
\end{align}

$\mathbf{I}$ in equation \ref{eq:stochastic_agg} corresponds to the input image, whereas $\tilde{\mathbf{I}}$ denotes the noisy input and $m$ denotes the binary mask. $\text{SM}_c(.)$ is the vanilla saliency map and $\tilde{\text{SM}}_c$ corresponds to the final aggregated saliency. 

The amount of perturbation for SmoothGrad is controlled by the standard deviation, $\sigma$, (also called noise level) of the Gaussian noise. Whereas for BinaryMask, the binary probability $p$ controls the perturbation magnitude. With a higher binary probability $p$, a higher number of input pixels are in $\tilde{\mathbf{I}}$, which means lower binary perturbation.  For our experiment, we fixed the noise level as $0.5$ and the binary probability $p$ as 0.90. $n=50$ samples have been selected from the neighborhood for our experiments.  
We  added these noises to the input images as additional augmentations during fine-tuning. ``\textit{Model-pert-binary}" and ``\textit{Model-pert-gaussian}" are two finetuned classifiers augmented by binary and Gaussian noise, respectively. 


\begin{figure*}[h]
\begin{center}
   \includegraphics[width=1.00\linewidth]{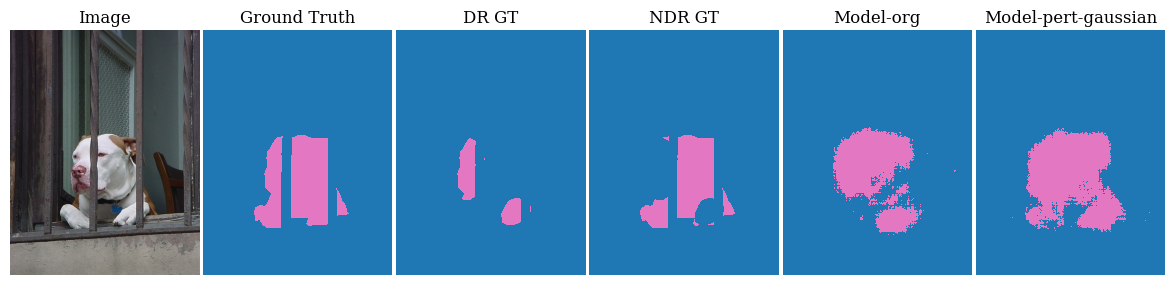}
   \includegraphics[width=1.00\linewidth]{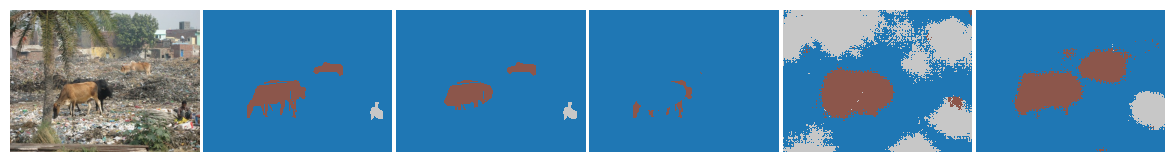}
   \includegraphics[width=1.00\linewidth]{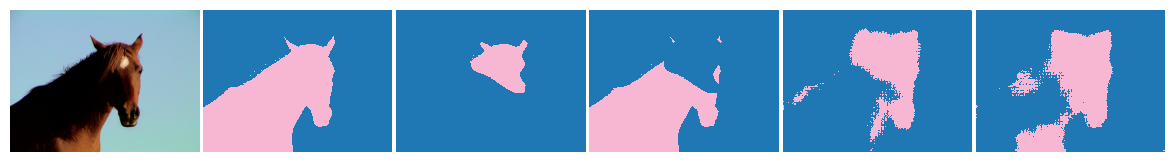}
   \includegraphics[width=1.00\linewidth]{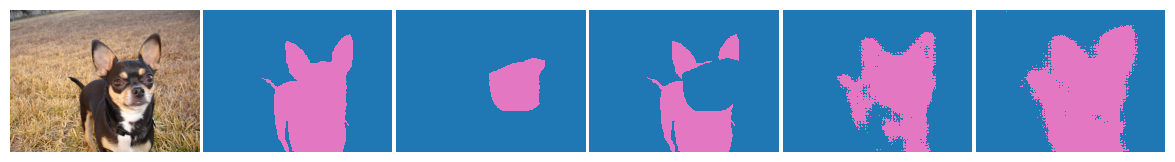}
   \includegraphics[width=1.00\linewidth]{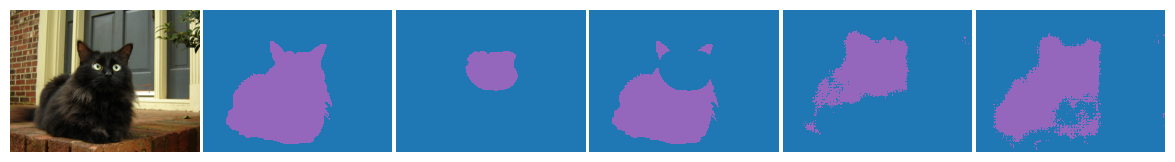}
   \includegraphics[width=1.00\linewidth]{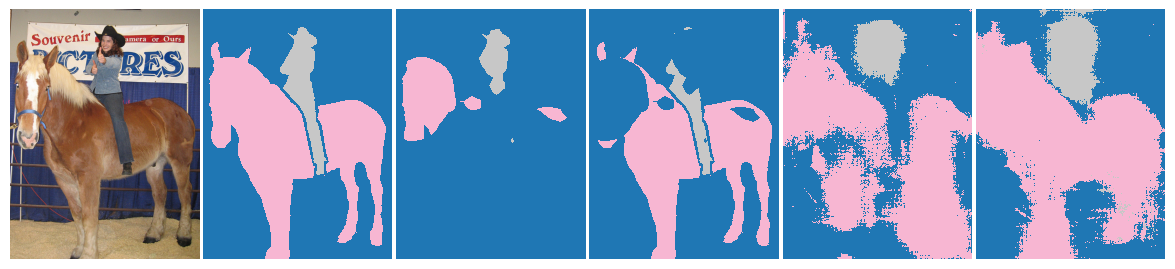}
   \end{center}
   \caption{Visual comparison of SmoothGrad saliencies between ``Model-org" and ``Model-pert-gaussian" fine-tuned model. Saliencies with basic background resolve are shown in the figure.}
\label{fig:supple_viz_compare_gaussian_noise}
\end{figure*}

\begin{figure*}[h]
\begin{center}
   \includegraphics[width=1.00\linewidth]{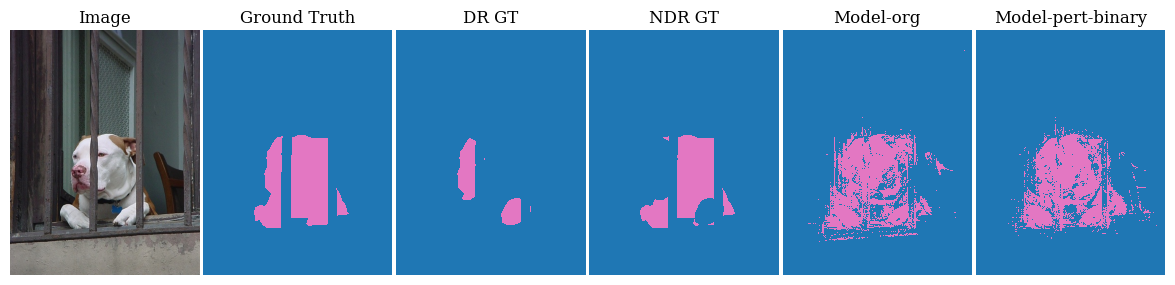}
   \includegraphics[width=1.00\linewidth]{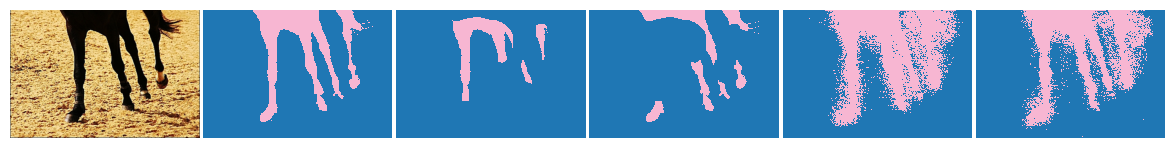}
   \includegraphics[width=1.00\linewidth]{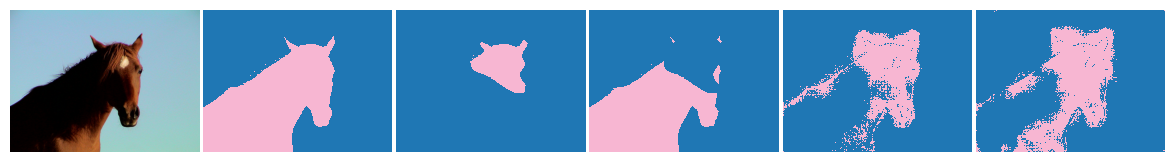}
   \includegraphics[width=1.00\linewidth]{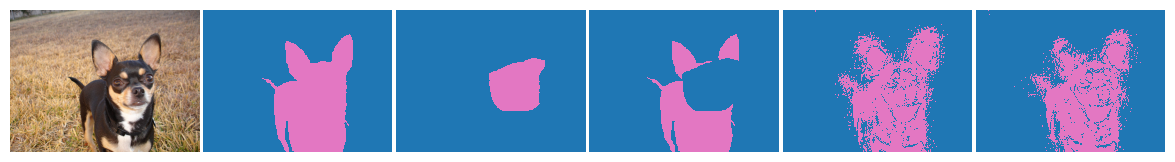}
   \includegraphics[width=1.00\linewidth]{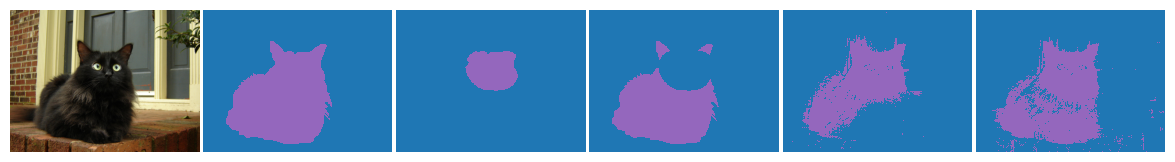}
   \end{center}
   \caption{Visual comparison of BinaryMask saliencies between ``Model-org" and ``Model-pert-binary" fine-tuned model. Saliencies with basic background resolve are shown in the figure.}
\label{fig:supple_viz_compare_binary_noise}
\end{figure*}

Figure \ref{fig:supple_viz_compare_gaussian_noise} presents a visual comparison of SmoothGrad saliencies derived from the ``Model-org" and ``Model-pert-gaussian" models. As SmoothGrad employs a stochastic aggregation approach, the basic background resolution yields significantly smoother saliencies for both models. Nevertheless, the ``Model-pert-gaussian" model exhibits superior saliencies in terms of the performance metrics discussed in Section \ref{sec:setup_eval_metrics}. In terms of visual quality, the ``Model-pert-gaussian" column of Figure \ref{fig:supple_viz_compare_gaussian_noise} exhibits superior saliencies compared to the ``Model-org" column. However, the perturbed model occasionally generates overly smooth saliencies, resulting in unclear object boundaries, as observed in the first and sixth rows of Figure \ref{fig:supple_viz_compare_gaussian_noise}. Similarly, Figure \ref{fig:supple_viz_compare_binary_noise} offers a visual comparison of BinaryMask saliencies for the ``Model-org" and ``Model-pert-binary" models. In this case, the ``Model-pert-binary" model demonstrates higher quality saliencies, as observed in the first, second, and sixth rows of Figure \ref{fig:supple_viz_compare_binary_noise}. Reinforcing the insights obtained from Section \ref{sec:stochastic_saliency}, both these figures support the notion that the classification model should be fine-tuned using similar noise in order to yield better-quality saliencies.

\subsection{Analysis of The Sensitivity Towards Noise.}
\begin{figure*}[t]
\begin{center}
   \includegraphics[width=0.39\linewidth]{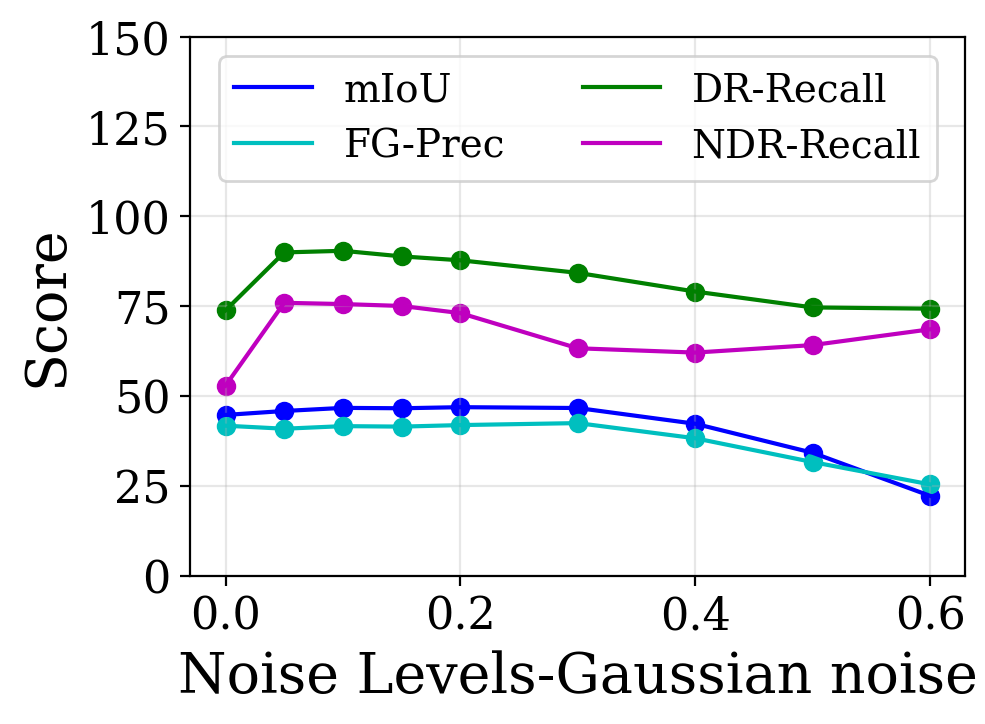}
   \includegraphics[width=0.39\linewidth]{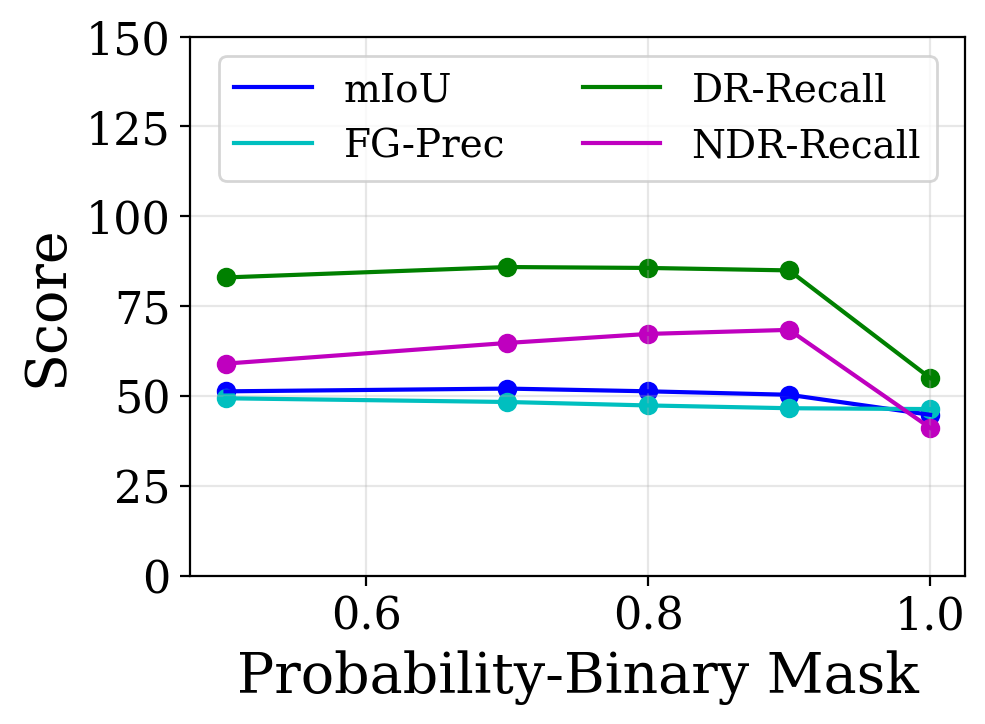}
\end{center}
   \caption{Sensitivity plots of the performance towards Gaussian noise levels $\sigma$ (left); towards binary probability $p$ (right).}
\label{fig:sensitivity_gaussian_binmask}
\end{figure*}

\begin{figure*}[t]
\begin{center}
   \includegraphics[width=0.39\linewidth]{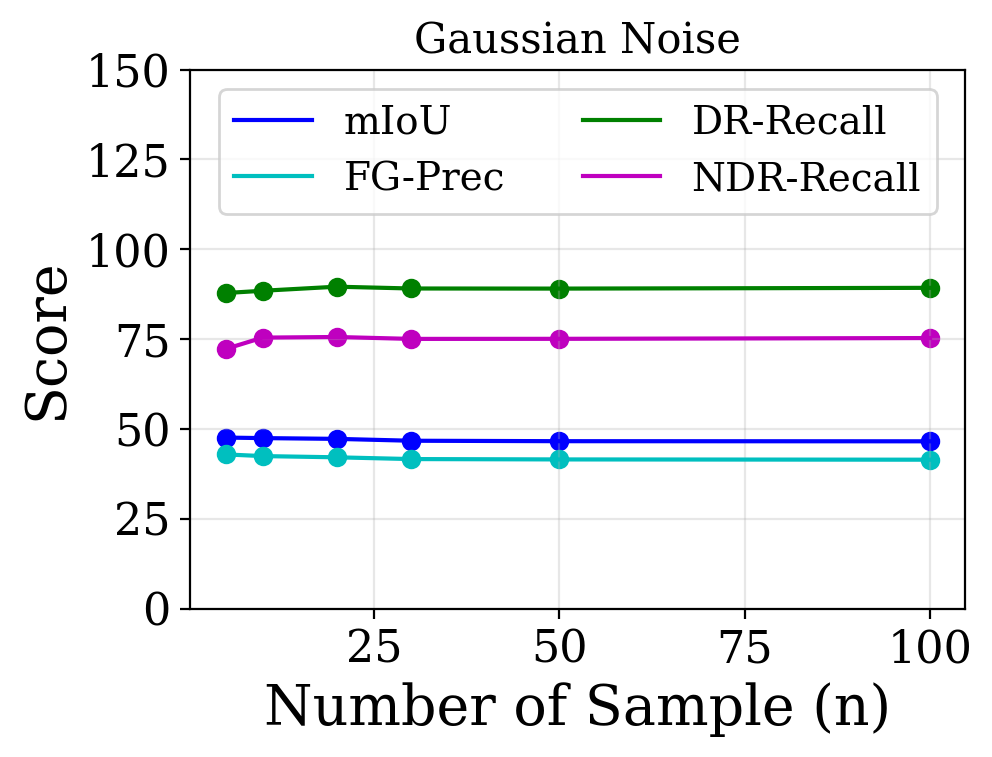}
   \includegraphics[width=0.39\linewidth]{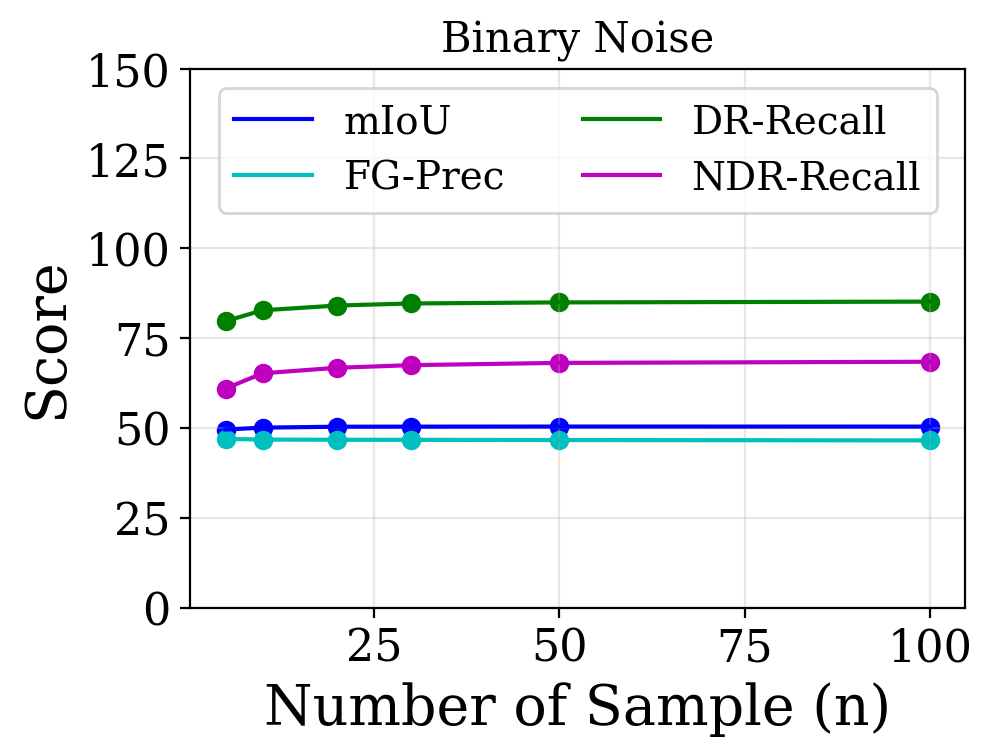}
\end{center}
   \caption{Sensitivity plots of the performance towards the number of samples $n$ for (left) SmoothGrad; (right) BinaryMask.}
\label{fig:sensitivity_num_samples_gaussian_binmask}
\end{figure*}

Figure \ref{fig:sensitivity_gaussian_binmask} and \ref{fig:sensitivity_num_samples_gaussian_binmask} illustrate the sensitivity of performance scores concerning the magnitude of noise and the number of neighborhood samples, respectively. In the case of SmoothGrad, the noise levels (standard deviation $\sigma$) dictate the magnitude of perturbation, with a higher $\sigma$ corresponding to a greater noise magnitude. Conversely, for BinaryMask, the binary probability $p$ governs the perturbation magnitude, with a lower probability $p$ corresponding to a higher level of perturbation. As evident from Figure \ref{fig:sensitivity_gaussian_binmask}, the models demonstrate sensitivity towards increased perturbation, with the NDR-Recall decreasing for higher noise levels in both cases. SmoothGrad exhibits greater sensitivity to higher perturbation, while BinaryMask displays less sensitivity to perturbation magnitude in terms of mIoU and FG-precision. As illustrated in Figure \ref{fig:sensitivity_num_samples_gaussian_binmask}, the performance remains relatively stable for the number of samples $n > 20$. However, when $n < 20$, the performance improves as the number of samples increases. 

\begin{figure*}[t]
\begin{center}
   \includegraphics[width=0.99\linewidth]{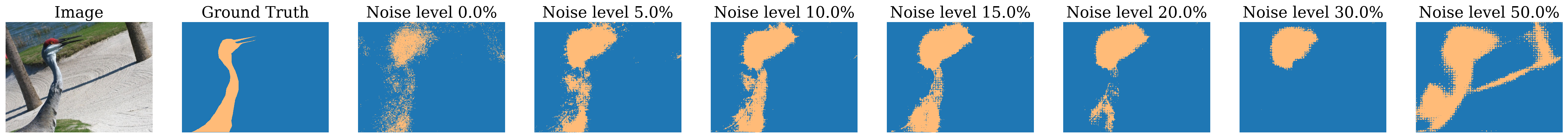}
   \includegraphics[width=0.99\linewidth]{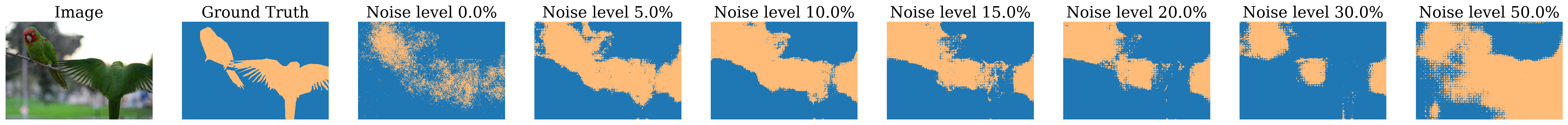}
   \includegraphics[width=0.99\linewidth]{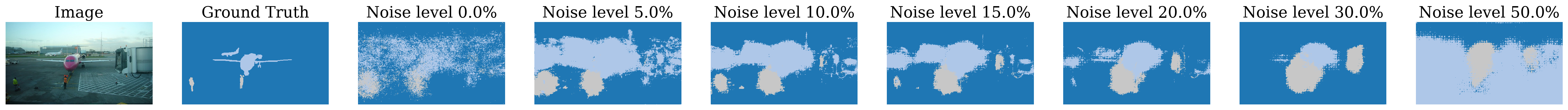}
\end{center}
   \caption{Visual evaluation of the sensitivity towards the noise level $\sigma$ of the Gaussian noise (SmoothGrad saliency with basic background resolve).}
\label{fig:viz_sensitivity_gaussian}
\end{figure*}

\begin{figure*}[t]
\begin{center}
   \includegraphics[width=0.99\linewidth]{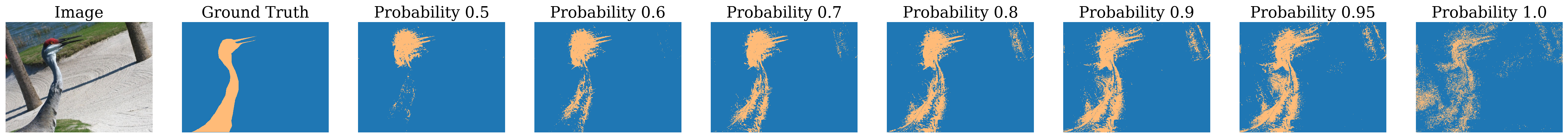}
   \includegraphics[width=0.99\linewidth]{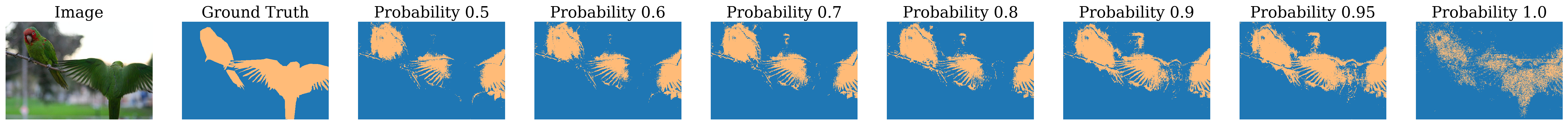}
   \includegraphics[width=0.99\linewidth]{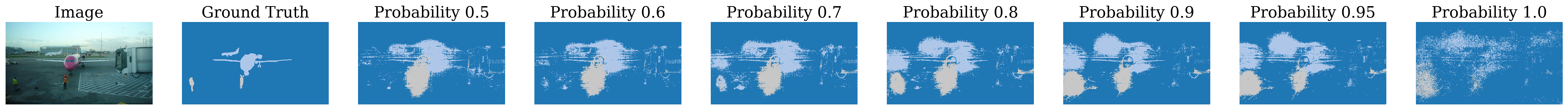}
\end{center}
   \caption{Visual evaluation of the sensitivity towards the binary probability of the perturbation (BinaryMask saliency with basic background resolve).}
\label{fig:viz_sensitivity_binmask}
\end{figure*}

By examining Figure \ref{fig:viz_sensitivity_gaussian}, we can see that excessively adding noise to the input image has a negative impact. As a result, the mIoU performance decreases for noise levels above $0.20$. Adding noise may make the saliency maps smoother; however, with increasing noise, the saliency maps may become unstable (shown in the noise level $50\%$ column). Figure \ref{fig:viz_sensitivity_binmask} depicts the sensitivity of BinaryMask saliency with respect to binary probability. The visualization reveals that as perturbation increases (low probability), the saliencies become less stable, as shown in the third and fourth columns of Figure \ref{fig:viz_sensitivity_binmask}. In contrast, higher probability leads to enhanced saliency quality, as evident in the seventh and eighth columns of Figure \ref{fig:viz_sensitivity_binmask}. It is important to note that a binary probability of $1.0$ does not involve any stochastic aggregation, as all pixels are selected to compute the saliency.

\section{Stochastic Aggregation Through Cropping}

\subsection{Analysis of The Sensitivity For Random Cropping.}

\begin{figure*}[h]
\begin{center}
   \includegraphics[width=0.40\linewidth]{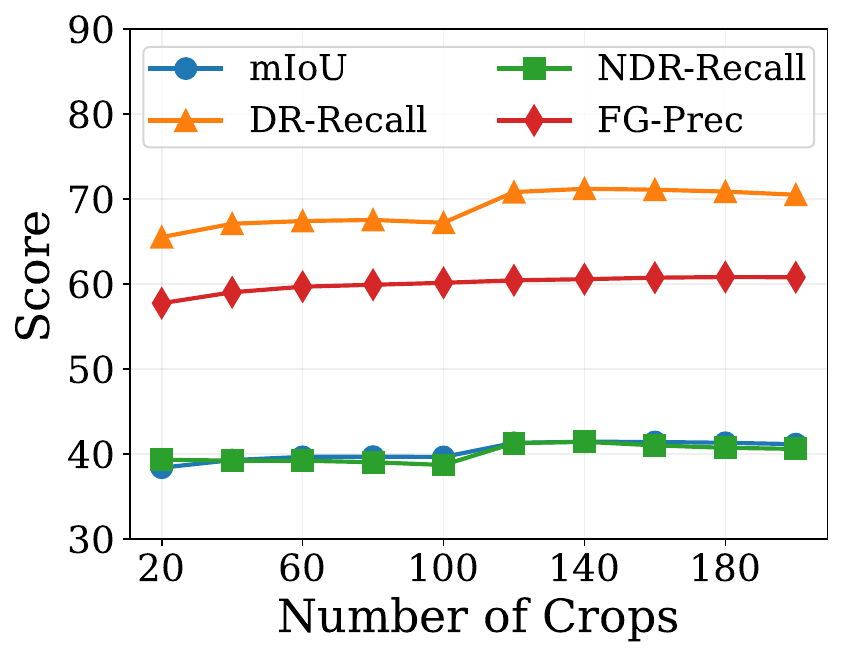}
   \includegraphics[width=0.40\linewidth]{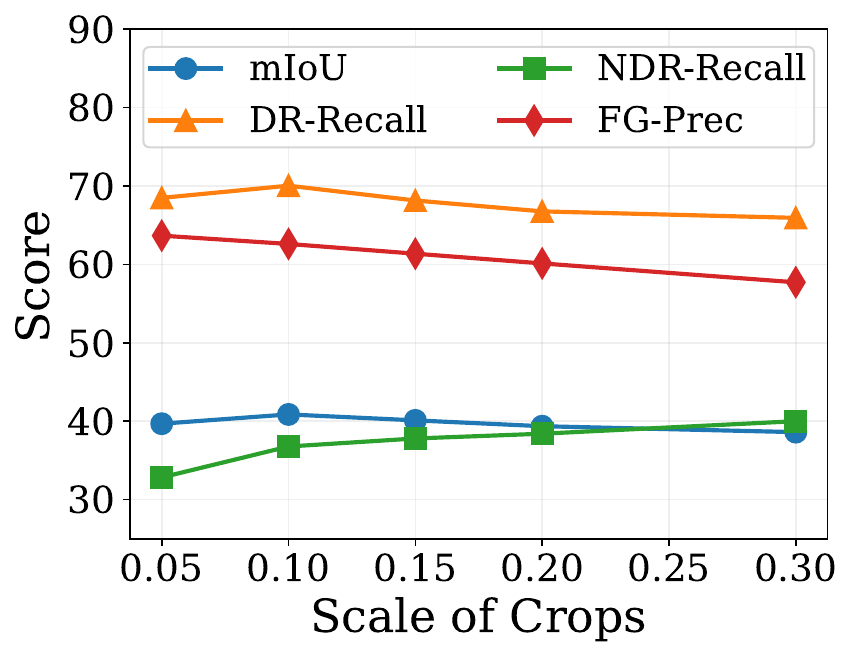}
\end{center}
   \caption{Sensitivity plots of the performance for random cropping (left) to the number of crops; (right) to the scale of the crops.}
\label{fig:sensitivity_random_crops}
\end{figure*}

Figure \ref{fig:sensitivity_random_crops} shows the sensitivity of the performance metrics towards the number of crops and the scale of crops for random cropping. With an increasing number of crops, the performance of random cropping-based saliencies improves. However, after 140 crops, we see the performance saturates. Choosing the correct scale of random crops is critical for better performance. The scale of the crops should not be lower than 0.10. The performance of the random cropping method saturates after a scale of 0.10.

\subsection{Different Variations of Cropping}
In this subsection, we explore different variations of random cropping and patching techniques that break the spatial structure of input images. Random patching is an erasure-based method similar to the idea of cutout \cite{devries2017improved} technique. We divide the full image into $16 \times 16$ grid-wise patches for random patching. Then we randomly mask out some of the patches with a Bernoulli probability of $0.1$, also called patching probability. The random patching idea is similar to the BinaryMask method in the sense that we are turning off some grid of pixels instead of individual pixels with a probability. Using stochastic aggregation of the

Given a CAM of an input image, we also explore the discriminative patching idea, where the patching probability is the complement of the CAM score $S^c_{cam}$ for each patch for the $c$-th class. It is important to mention that $S^c_{cam}$ corresponds to the maximum CAM score across all the $C$ classes in the patch (where $C$ is the total number of classes). The discriminative patch (disc-Patch) is implemented as follows:
\begin{align}
    &p = \alpha * S_{cam}^c  \nonumber\\
    & \bar{m} = 1 - m; \quad m \sim \text{Bernoulli}(p) \nonumber\\
    &\tilde{\mathbf{I}} = \mathbf{I} \odot \bar{m} \nonumber
\end{align}
 $\bar{m}$ is the binary filter applied to the patches and $\tilde{\mathbf{I}}$ denotes the perturbed image. $S^c_{cam}$ is multiplied by $\alpha \in (0, 1)$ so that the discriminative patch probability does not reach $0$ for the most discriminative region. For our experiments, we choose $\alpha = 0.4$.

Similar to discriminative patching, we explore discriminative cropping, where the selection of each crop has a probability that is the complement to the CAM score $S^c_{cam}$ for that crop. The discriminative cropping is implemented as follows:
\begin{align}
    &p = \text{ReLU}(\beta - S^c_{cam}) \nonumber\\
    &\tilde{\text{SM}}_c(x) = \frac{1}{n} \sum_{i=1}^{n} m*w_i*\text{SM}_c(\tilde{\mathbf{I}}_i); \nonumber \\
    &m \sim \text{Bernoulli}(p) \nonumber
\end{align}

$\tilde{\text{SM}}_c(x)$ is the final aggregated saliency using discriminative cropping. $m$ is the binary filter applied to the crops and $\tilde{\mathbf{I}}$ denotes the perturbed image. We choose $\beta = 0.7$ for our experiments. 

\begin{figure*}[h]
\begin{center}
   \includegraphics[width=0.8\linewidth]{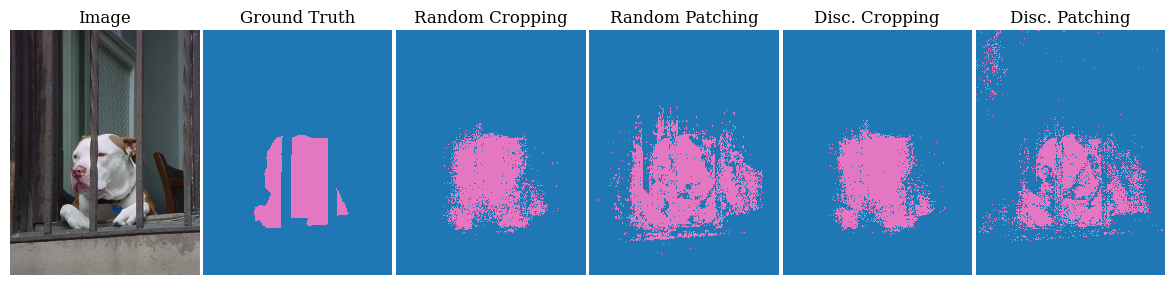}
   \includegraphics[width=0.8\linewidth]{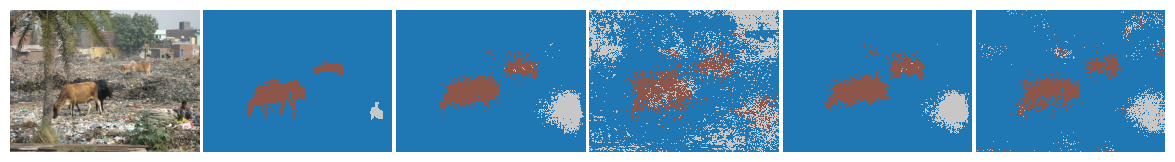}
   \includegraphics[width=0.8\linewidth]{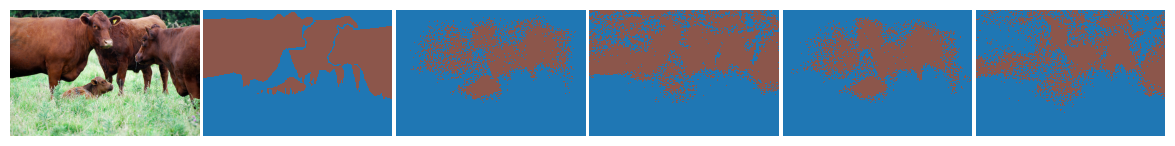}
   \includegraphics[width=0.8\linewidth]{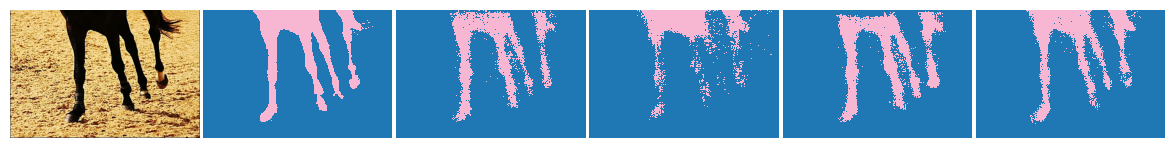}
   \includegraphics[width=0.8\linewidth]{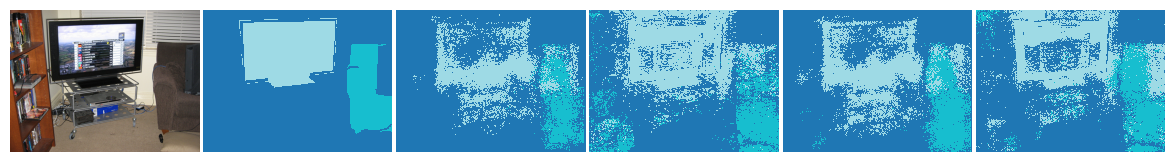}
   \includegraphics[width=0.8\linewidth]{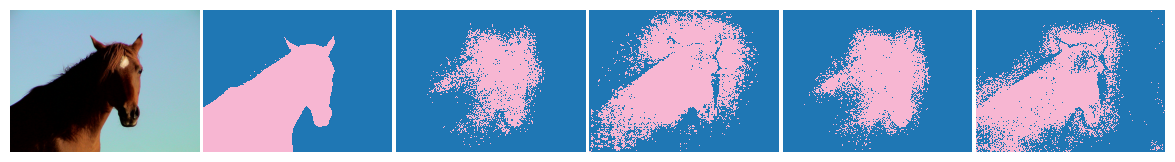}
   \includegraphics[width=0.8\linewidth]{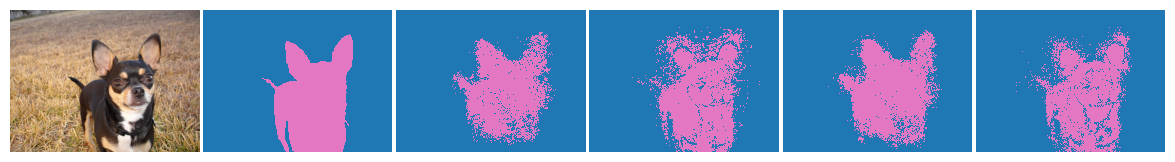}
   \includegraphics[width=0.8\linewidth]{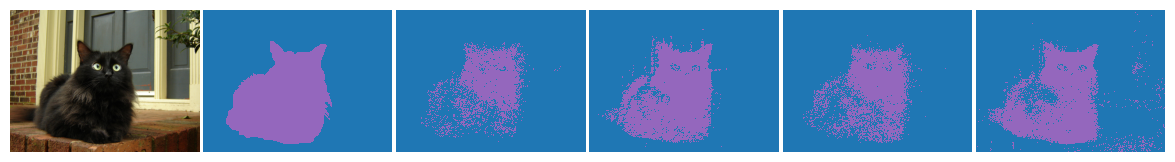}
   \includegraphics[width=0.8\linewidth]{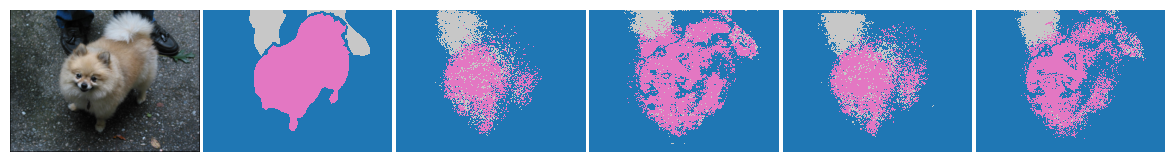}
   \includegraphics[width=0.8\linewidth, height=72px]{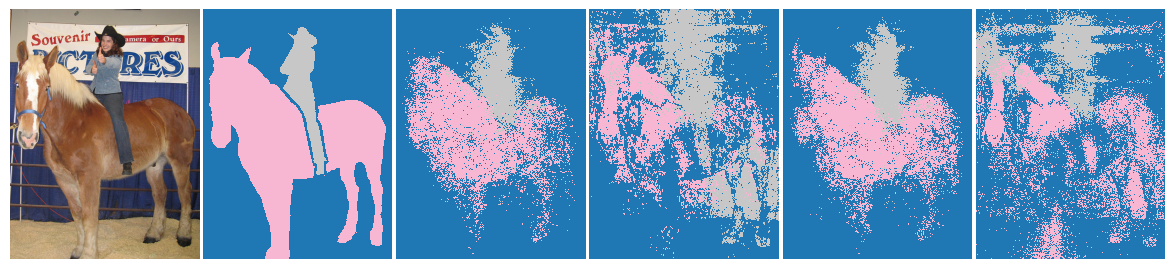}
\end{center}
   \caption{Visual comparison between Random Cropping, Random Patching, Discriminative Cropping, and Discriminative Patching saliencies. Saliencies with basic background resolve are shown in the figure.}
\label{fig:supple_viz_compare_cropping_patching}
\end{figure*}

Figure \ref{fig:supple_viz_compare_cropping_patching} provides a visual comparison of saliencies generated by Random Cropping, Random Patching, Discriminative Cropping, and Discriminative Patching. As discussed in Section \ref{sec:stochastic_random_crop}, both Random Cropping and Discriminative Cropping display higher quality and more stable saliencies. In contrast, the saliencies produced by Random Patching and Discriminative Patching are less stable, primarily due to the fact that the classification model has not been fine-tuned with similar noise perturbation. For instance, in Figure \ref{fig:supple_viz_compare_cropping_patching}, the second and fourth rows display poor saliency maps for the patching methods. Conversely, the sixth, seventh, and eighth rows exhibit higher-quality saliencies for the patching method. Moreover, the discriminative variations of both these methods demonstrate a modest enhancement in saliency quality, as evidenced by the second, fourth, sixth, and seventh rows of Figure \ref{fig:supple_viz_compare_cropping_patching}.

\end{document}